\documentclass[a4paper]{article}

\usepackage[english]{babel}
\usepackage[utf8x]{inputenc}
\usepackage[T1]{fontenc}
\usepackage{booktabs}
\usepackage{subfig}
\usepackage[titletoc,title]{appendix}
\usepackage{amsfonts}
\usepackage{float}
\usepackage{array}
\usepackage{multirow}

\usepackage[a4paper,top=3cm,bottom=2cm,left=3cm,right=3cm,marginparwidth=1.75cm]{geometry}

\usepackage{amsmath}
\usepackage{graphicx}
\usepackage[colorinlistoftodos]{todonotes}
\usepackage{authblk}

\title{Neural-net-induced Gaussian process regression for function approximation and PDE solution}

\date{}

\author[]{Guofei Pang}
\author[]{Liu Yang}
\author[]{George Em Karniadakis\thanks{Corresponding author: george\_karniadakis@brown.edu}}
\affil[]{Division of Applied Mathematics, Brown University, Providence, RI 02912, USA}

\begin{document}
\maketitle

\begin{abstract}
Neural-net-induced Gaussian process (NNGP) regression inherits both the high expressivity of deep neural networks (deep NNs) as well as the uncertainty quantification property of Gaussian processes (GPs). We generalize the current NNGP to first include a larger number of hyperparameters and subsequently train the model by maximum likelihood estimation. Unlike previous works on NNGP that targeted classification, here we apply the generalized NNGP to function approximation and to solving partial differential equations (PDEs). Specifically, we develop an analytical iteration formula to compute the covariance function of GP induced by deep NN with an error-function nonlinearity. We compare the performance of the generalized NNGP for function approximations and PDE solutions with those of GPs and fully-connected NNs. We observe that for smooth functions the generalized NNGP can yield the same order of accuracy with GP, while both NNGP and GP outperform deep NN. For non-smooth functions, the generalized NNGP is superior to GP and comparable or superior to deep NN.

\end{abstract}

\section{Introduction}
The high expressivity of deep NNs \cite{poole2016exponential} can be combined with the predictive ability of GP regression by taking advantage of the apparent equivalence between GPs and fully-connected, infinitely-wide, deep NNs with random weights and biases. This equivalence was first proved by \cite{neal1995bayesian} for shallow NNs. In a subsequent paper, the authors of \cite{williams1997computing} obtained analytically the covariance functions of the GPs induced by the shallow NNs having two specific activation functions: error and Gaussian functions. More recently, the authors of \cite{lee2017deep} extended \cite{williams1997computing}'s work to the case of deep NNs having generic activation functions, with reference to the mean field theory of signal propagation \cite{poole2016exponential,sompolinsky1988chaos}. 

The term NNGP was coined by the authors of \cite{lee2017deep}, but until now this method despite its potential merits has not been used extensively in applications. In particular, the authors of \cite{lee2017deep} validated the NNGP on two image recognition datasets: MNIST (handwritten digits data) and CIFAR-10 (color images data), and observed that NNGP outperforms finite width NNs in terms of classification accuracy. The primary objective of our work is to exploit the flexibility of NNGP in order to approximate smooth and discontinuous functions, and furthermore in assessing the effectiveness of NNGP to solve linear and nonlinear PDEs, compared to the GP and deep NN methods formulated recently in \cite{raissi2017inferring,raissi2017numerical,raissi2017physics}. Specifically, so far GP regression has been successfully applied to solution of PDEs including linear and nonlinear forward problems \cite{raissi2017inferring}  as well as inverse problems \cite{pang2017discovering,raissi2017machine}. Use of GP regression can bypass discretizing differential operators by properly placing GP priors, and can easily handle noisy data as well as the associated uncertainty propagation in time-dependent problems. Hence, the motivation of the current research is to evaluate if indeed NNGP possesses not only the high expressivity of deep NNs but also the uncertainty quantification property of GPs. To this end, we will perform three-way comparisons between NNGP, GP and deep NN for prototypical problems involving function approximation and PDE solution.  

The present paper makes two main contributions. 

First, to pursue high expressivity, we increase the number of hyperparameters of NNGP by assuming that the prior variances of weights and biases vary layer by layer and that the weight variances between the input layer and the first hidden layer are different for different neurons in the input layer. We also train the resulting NNGP regression model by using maximum likelihood estimation. We note that the authors of \cite{lee2017deep} assumed only two hyperparameters: one weight variance and one bias variance, which are kept fixed over different layers; the two hyperparameters are adjusted to achieve the best performance on validation dataset. However, no training was performed in \cite{lee2017deep}. 

Second, we derive the analytical covariance function for the NNGP induced by the deep NN with the error-function nonlinearity. This is motivated by the observation that the analytical covariance function induced by ReLU activated NN, given in \cite{lee2017deep}, cannot be used to solve the two-dimensional Poisson equation. Denoting by $k(\boldsymbol{x},\boldsymbol{x}')$ the analytical covariance function from the ReLU case where $\boldsymbol{x}$ and $\boldsymbol{x}'$ are input vectors of length two, we observed that the Laplacian of the covariance function, namely, $\Delta_{\boldsymbol{x}}\Delta_{\boldsymbol{x}'}k(\boldsymbol{x},\boldsymbol{x}')$ , which has to be computed in order to solve the problem (see Eq.(\ref{kernel_f})), will tend to infinity as $\boldsymbol{x}\rightarrow \boldsymbol{x}'$.

The paper is organized as follows. In Section \ref{equivalence_sec}, we review the equivalence between GPs and infinitely wide NNs, which is a cornerstone of NNGP. We change the notations for weight and bias variances compared to the previous work \cite{lee2017deep} in order to show a different hyperparameter setup. Section \ref{Two-kernel-sec} introduces the analytically tractable covariance functions induced by the deep NNs with the ReLU and error-function nonlinearities. We employed a version of Gaussian quadrature, proposed in \cite{lee2017deep}, to validate numerically the two analytical covariance functions. The methodology on GP regression for function approximation and PDE solution is briefly reviewed and summarized in Section \ref{GPR-sec}. Section \ref{numerical-sec} compares the accuracy of GP, NNGP and NN in function approximation and PDE solution. The uncertainty estimates of GP and NNGP are also presented in the section. Finally, we conclude in Section \ref{numerical-sec} with a brief summary.

\begin{figure}[H] 
\centering
\includegraphics[width=.9\textwidth]{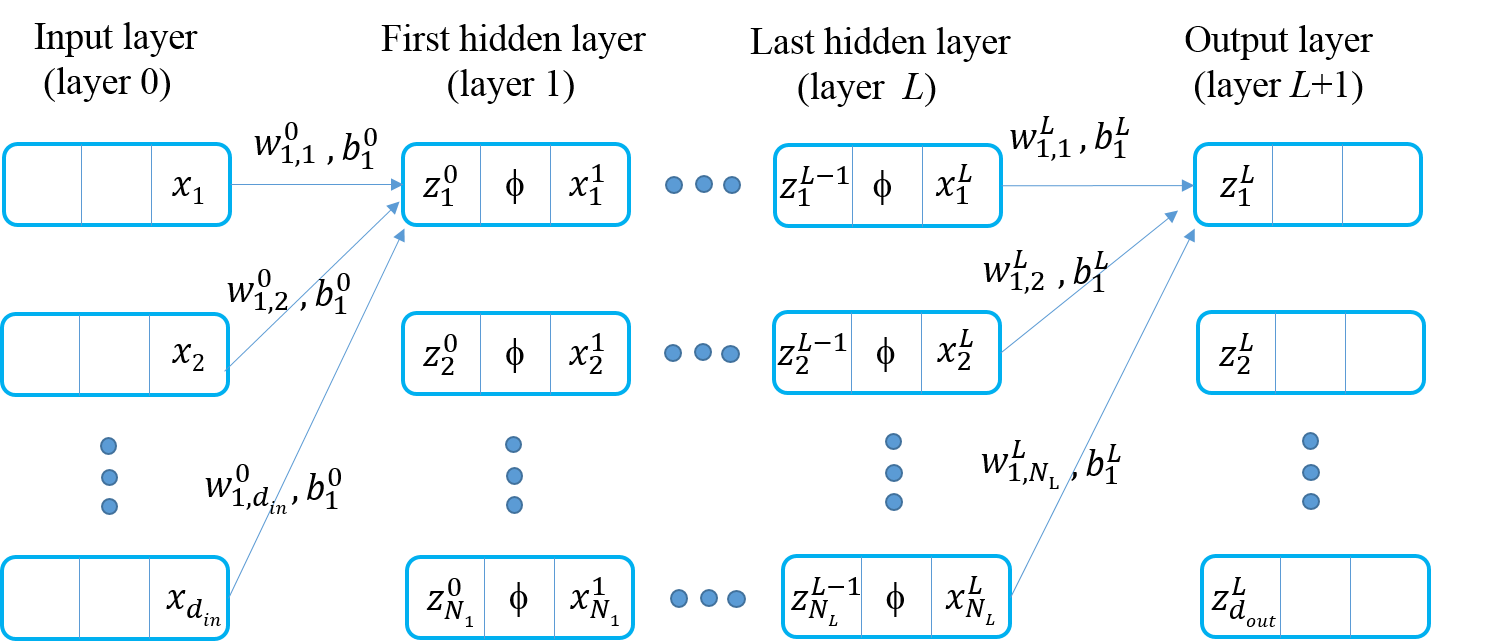}
\caption{\label{illustrating_NN} A fully connected neural net with $L$ hidden layers. For each unit or neuron in hidden layers, there exist one input $z^{l-1}_i$ and one output $x^l_i$ with $l=1,1\cdots,L$. Layer 0 is the input layer. $x_i$ is the $i$-th component of the input vector $\boldsymbol{x}$, and $z^L_i$ is the $i$-th component of the output vector $\boldsymbol{z}$. The dimensions of input and output spaces are $d_{in}$ and $d_{out}$, respectively. The width for hidden layer $l$ is $N_l$. At the center of each unit is the activation function $\phi(\cdot)$ that transforms input to the corresponding output. Between two successive layers, the weight $w^{l}_{ij}$ for $l=0,1,\cdots,L$ denotes the contribution of unit $j$ in layer $l$ to unit $i$ in layer $l+1$. Layer $L+1$ is the output layer. The bias $b^{l}_i$ is attached to unit $i$ in layer $l+1$ for $l=0,1,\cdots,L$. Note that for clarity most of connecting arrows between layers are omitted. }
\end{figure}

\section{Equivalence between GPs and infinitely wide NNs} \label{equivalence_sec}

In this section, we adopt nearly the same notations as \cite{lee2017deep} except those for weight and bias variances. We add the subscript $l$ to the variances $\sigma^2_w$ and $\sigma^2_b$ in order to represent index of layer and the subscript $j$ to $\sigma^2_w$ in order to distinguish the weights coming from different neurons in input layer.

The fully-connected neural net can be represented as in Fig.\ref{illustrating_NN}. The equivalence relies on two assumptions: (1) Weight and bias parameters are all random variables. For $l \ge 1$ the weights $w^{l}_{ij}$ are independent and identically distributed (i.i.d.) and obey the same distribution $\mathcal{D}_l$ with zero mean and same variance, namely, $w^{l}_{ij}\sim \mathcal{D}_l(0,\sigma^2_{w,l}/N_l), \forall i,j$.  The biases $b^{l}_i$ ($l \ge 1$) are i.i.d. Gaussian random variables, $b^{l}_i \sim \mathcal{N}(0,\sigma^2_{b, l}), \forall i$. The weights and biases are independent with each other. Note that in \cite{lee2017deep} the weight and bias variances are kept fixed over different layers, i.e., $\sigma^2_{w,l}\equiv\sigma^2_w$ and $\sigma^2_{b,l}\equiv\sigma^2_b$, but we here assume the variances vary layer by layer and increase the number of variances to be determined. For $l=0$, the weights need only to be independent with respect to the subscript $i$, i.e., $w^{0}_{ij}\sim \mathcal{D}_0(0, \sigma^2_{w,0,j}), \forall i$; the biases are independent Gaussian random variables, $b^0_i \sim \mathcal{N}(0, \sigma^2_{b,0}), \forall i$. Unlike \cite{lee2017deep}, we remove the assumption of being identically distributed for the weights. These weights do not have to be identically distributed since we cannot use the Central Limit Theorem for the input of the first hidden layer, namely, $z^0_i$. Actually, the width of input layer is finite and will never be infinity. Allowing weight variances to vary for different neurons in input layer, we increase again the number of variances to be determined in NNGP. (2) The widths of all hidden layers tend to infinity, i.e., $N_l\rightarrow \infty, 1\le l \le L$. This allows one to use the Central Limit Theorem to deduce the equivalence relation.

It follows from \cite{williams1997computing} and \cite{cho2009kernel} that for analytical derivation of the covariance function the distribution $\mathcal{D}_0$ must be Gaussian. The specific forms of $D_l, l \ge 1$ do not really matter, since using the Central Limit Theorem one always obtains the Gaussian distribution whatever $\mathcal{D}_l$ is.

The $i$-th component of the input of the second hidden layer is computed as

\begin{equation}
z^1_i(\boldsymbol{x}) =\sum_{j=1}^{N_1}w^1_{ij}x^1_j(\boldsymbol{x})+b^1_i, \quad x^1_j(\boldsymbol{x})=\phi\left(\sum_{k=1}^{d_{in}}w^0_{jk}x_k+b^0_j\right).
\end{equation}
In the rest of paper, we use bold letter, say $\boldsymbol{x}$ and $\boldsymbol{K}$, to represent column vector or matrix. The dependence on the input vector $\boldsymbol{x}$ is emphasized in the above equations. Since weights and biases $w^0_{jk}$ and $b^0_j$ are i.i.d., the output of the first hidden layer $x^1_j(\boldsymbol{x})$ is also i.i.d. with respect to the subscript $j$. Because $z^1_i(\boldsymbol{x})$ is a sum of i.i.d. terms, the use of the Central Limit Theorem yields that $z^1_i(\boldsymbol{x})$ will be Gaussian distributed if $N_1\rightarrow \infty$. Similarly, from the multidimensional Central Limit Theorem, any finite collection of $\{z^1_i(\boldsymbol{x}_1),\cdots,z^1_i(\boldsymbol{x}_k)\}$ will form a joint multivariate Gaussian distribution, which is exactly the definition of a Gaussian process. Note that $\boldsymbol{x}_1,\cdots,\boldsymbol{x}_k$ are $k$ arbitrary input vectors. 

Denoting by $\mathcal{GP}(\mu(\boldsymbol{x}),k(\boldsymbol{x},\boldsymbol{x}'))$ the GP with the mean function $\mu(\cdot)$ and the covariance function $k(\cdot,\cdot)$, we learn that $z^1_i(\boldsymbol{x})\sim \mathcal{GP}(\mu^1(\boldsymbol{x}),k^1(\boldsymbol{x},\boldsymbol{x}'))$. Also, $\{z^1_1(\boldsymbol{x}),z^1_2(\boldsymbol{x}),\cdots,z^1_{N_2}(\boldsymbol{x})\}$ are independent GPs with same mean and covariance functions. The mean function of GP $z^1_i(\boldsymbol{x})$ is zero due to the zero mean of the distribution $\mathcal{D}_1$, and the covariance function is computed as 
\begin{equation}\label{cov_2_layer}
\begin{split}
k^1(\boldsymbol{x},\boldsymbol{x}')=\mathbb{E}(z^1_i(\boldsymbol{x})z^1_i(\boldsymbol{x}')) & =\mathbb{E}\left(\left(\sum_{j=1}^{N_1}w^{1}_{ij}x^1_j(\boldsymbol{x})+b^1_i\right)\left(\sum_{k=1}^{N_1}w^{1}_{ik}x^1_k(\boldsymbol{x}')+b^1_i\right)\right) \\
& = \mathbb{E}\left(\left(\sum_{j=1}^{N_1}w^{1}_{ij}\phi(z^0_j(\boldsymbol{x}))+b^1_i\right)\left(\sum_{k=1}^{N_1}w^{1}_{ik}\phi(z^0_k(\boldsymbol{x}'))+b^1_i\right)\right) \\
&=\sum_{j=1}^{N_1}\mathbb{E}(w^{1}_{ij})^2\mathbb{E}\left[\phi(z^0_j(\boldsymbol{x}))\phi(z^0_j(\boldsymbol{x}'))\right]+\mathbb{E}(b^1_i)^2 \\
&=\frac{\sigma^2_{w,1}}{N_1}\sum_{j=1}^{N_1}\mathbb{E}\left[\phi(z^0_j(\boldsymbol{x}))\phi(z^0_j(\boldsymbol{x}'))\right]+\sigma^2_{b,1}\\
&=\sigma^2_{w,1}\mathbb{E}\left[\phi(z^0_j(\boldsymbol{x}))\phi(z^0_j(\boldsymbol{x}'))\right]+\sigma^2_{b,1}.\\
\end{split} 
\end{equation}
In the above derivation, we use the independence relations between weights $\mathbb{E}(w^{1}_{ij}w^{1}_{ik})=\delta_{jk}\sigma^2_{w,1}$ ($\delta_{ij}$ is the Kronecker delta), between weight and bias $\mathbb{E}(w^{1}_{ij}b^1_i)=\mathbb{E}(w^{1})\mathbb{E}(b^1_i)=0$, and between weight and activation function $\mathbb{E}\left[(w^1_{ij})^2\phi(z^0_j(\boldsymbol{x}))\phi(z^0_j(\boldsymbol{x}'))\right]=\mathbb{E}\left(w^1_{ij}\right)^2\mathbb{E}\left[\phi(z^0_j(\boldsymbol{x}))\phi(z^0_j(\boldsymbol{x}'))\right]$. The last equality in Eq. (\ref{cov_2_layer}) comes from the independence of $\phi(z^0_j(\boldsymbol{x}))\phi(z^0_j(\boldsymbol{x}'))$ with respect to $j$. 
The input of the first hidden layer is denoted by $z^0_j(\boldsymbol{x})=\sum_{k=1}^{d_{in}}w^0_{jk}x_k+b^0_j$, and the expectation $\mathbb{E}(\cdot)$ is taken over the distribution of $w^0_{ij}$ and $b^0_i$. Actually, $z^0_j(\boldsymbol{x})$ is a GP only if the distribution $D_0$ is Gaussian. In this case, the covariance function of the GP $z^0_j(\boldsymbol{x})$ is
\begin{equation}\label{cov_1_layer}
\begin{split}
k^0(\boldsymbol{x},\boldsymbol{x}')=\mathbb{E}(z^0_i(\boldsymbol{x})z^0_i(\boldsymbol{x}')) & =\mathbb{E}\left(\left(\sum_{j=1}^{d_{in}}w^{0}_{ij}x_j+b^0_i\right)\left(\sum_{k=1}^{d_{in}}w^{0}_{ik}x'_k+b^0_i\right)\right) \\
&=\sum_{j=1}^{d_{in}}\mathbb{E}(w^{0}_{ij})^2x_j x'_j+\mathbb{E}(b^0_i)^2 \\
&=\sum_{j=1}^{d_{in}}\sigma^2_{w,0,j}x_j x'_j+\sigma^2_{b,0}.
\end{split} 
\end{equation}
The input vector components $x_j$ and $x'_j$ are deterministic and thus $\mathbb{E}(x_j x'_j)=x_j x'_j.$

\section{Two NN-induced covariance kernels of GP}\label{Two-kernel-sec}
The arguments of the previous section can be extended to deeper layers by induction. For layer $l$ ($\ge 2$), using the multidimensional Central Limit Theorem, we learn that $z^l_i(\boldsymbol{x})$ and $z^{l-1}_i(\boldsymbol{x})$ are both GPs. Specifically, $z^l_i(\boldsymbol{x})$ has zero mean function and the covariance function computed as 
\begin{equation}\label{deeper_layer}
\begin{split}
k^l(\boldsymbol{x},\boldsymbol{x}')=\mathbb{E}(z^{l}_i(\boldsymbol{x})z^{l}_i(\boldsymbol{x}'))  
&=\sigma^2_{w,l}\mathbb{E}\left[\phi(z^{l-1}_i(\boldsymbol{x}))\phi(z^{l-1}_i(\boldsymbol{x}'))\right]+\sigma^2_{b, l}.
\end{split} 
\end{equation}
The expectation in Eq. (\ref{deeper_layer}) is taken over the GP governing $z^{l-1}_i(\boldsymbol{x})$, and this amounts to integrating over the joint distribution of two Gaussian random variables $z^{l-1}_i(\boldsymbol{x})$ and $z^{l-1}_i(\boldsymbol{x}')$ given $\boldsymbol{x}$ and $\boldsymbol{x}'$. The covariance function is rewritten as
\begin{equation}\label{kl_E}
\begin{split}
k^l(\boldsymbol{x},\boldsymbol{x}')&=\sigma^2_{w,l}\int_{\mathbb{R}^2}\phi(z^{l-1}_i(\boldsymbol{x}))\phi(z^{l-1}_i(\boldsymbol{x}'))p(z^{l-1}_i(\boldsymbol{x}),z^{l-1}_i(\boldsymbol{x}'))d z^{l-1}_i(\boldsymbol{x})d z^{l-1}_i(\boldsymbol{x}')+\sigma^2_{b, l},
\end{split} 
\end{equation}
where the density function of the joint distribution of $z^l_i(\boldsymbol{x})$ and $z^l_i(\boldsymbol{x}')$ is 
\begin{equation}
p(z^{l-1}_i(\boldsymbol{x}),z^{l-1}_i(\boldsymbol{x}'))=\frac{1}{2\pi|\Sigma|^{\frac{1}{2}}}\exp\left(-\frac{1}{2}\left[\begin{array}{c}z^{l-1}_i(\boldsymbol{x})\\z^{l-1}_i(\boldsymbol{x}')\end{array}\right]^T\Sigma^{-1}\left[\begin{array}{c}z^{l-1}_i(\boldsymbol{x})\\z^{l-1}_i(\boldsymbol{x}')\end{array}\right]\right)
\end{equation} 
with the covariance matrix
\begin{equation}
\Sigma=\left[\begin{array}{cc}\mathbb{E}(z^{l-1}_i(\boldsymbol{x}))^2 & \mathbb{E}(z^{l-1}_i(\boldsymbol{x})z^{l-1}_i(\boldsymbol{x}'))\\ \mathbb{E}(z^{l-1}_i(\boldsymbol{x}')z^{l-1}_i(\boldsymbol{x})) & \mathbb{E}(z^{l-1}_i(\boldsymbol{x}'))^2\end{array}\right]=
\left[\begin{array}{cc}k^{l-1}(\boldsymbol{x},\boldsymbol{x}') & k^{l-1}(\boldsymbol{x},\boldsymbol{x}')\\ k^{l-1}(\boldsymbol{x}',\boldsymbol{x}) & k^{l-1}(\boldsymbol{x}',\boldsymbol{x}')\end{array}\right].
\end{equation}
Eq. (\ref{kl_E}) is actually an iteration formula, given by
\begin{equation}
\begin{split}
k^{l}(\boldsymbol{x},\boldsymbol{x}')&=\sigma^2_{w,l}F_{\phi}\left(k^{l-1}(\boldsymbol{x},\boldsymbol{x}),k^{l-1}(\boldsymbol{x}',\boldsymbol{x}'),k^{l-1}(\boldsymbol{x},\boldsymbol{x}')\right)+\sigma^2_{b, l},l=1,2,\cdots,L,\\
k^0(\boldsymbol{x},\boldsymbol{x}')&=\boldsymbol{x}^T\boldsymbol{\Lambda}\boldsymbol{x}'+\sigma^2_{b,0},
\end{split}
\end{equation}
where the diagonal matrix $\boldsymbol{\Lambda}$ is defined by $\boldsymbol{\Lambda}=diag(\sigma^2_{w,0,1},\sigma^2_{w,0,2},\cdots,\sigma^2_{w,0,d_{in}})$. Note that $k^{l-1}(\boldsymbol{x},\boldsymbol{x}')=k^{l-1}(\boldsymbol{x}',\boldsymbol{x})$ due to symmetry of the kernel. The trivariate function $F_{\phi}()$ represents the two-fold integral in Eq. (\ref{kl_E}), which depends on specific form of activation function $\phi(\cdot)$. For certain $\phi(\cdot)$, the function $F_{\phi}()$ is analytically tractable. Two iteration formulas that are analytically derivable will be given in the next two subsections. 

\subsection{Kernel from the ReLU nonlinearity}
Letting the activation function be a rectifier $\phi(x)=\max(x,0)$, we can derive the analytical expression for $F_{\phi}()$, and the corresponding iteration formula is \cite{lee2017deep,cho2009kernel} 
\begin{equation}\label{iter_ReLU}
\begin{split}
k^{l}(\boldsymbol{x},\boldsymbol{x}')&=\frac{\sigma^2_{w,l}}{2\pi}\sqrt{k^{l-1}(\boldsymbol{x},\boldsymbol{x})k^{l-1}(\boldsymbol{x}',\boldsymbol{x}')}\left(\sin\theta^{l-1}+(\pi-\theta^{l-1})\cos\theta^{l-1}\right)+\sigma^2_{b, l},l=1,2,\cdots,L,\\
\theta^{l-1}&=\cos^{-1}\left(\frac{k^{l-1}(\boldsymbol{x},\boldsymbol{x}')}{\sqrt{k^{l-1}(\boldsymbol{x},\boldsymbol{x})k^{l-1}(\boldsymbol{x}',\boldsymbol{x}')}}\right),\\
k^0(\boldsymbol{x},\boldsymbol{x}')&=\boldsymbol{x}^T\Lambda\boldsymbol{x}'+\sigma^2_{b,0}.
\end{split}
\end{equation}
Here we have increased the number of undetermined variances compared to the original formula in \cite{lee2017deep} in which $\sigma^2_{w,l}=\sigma^2_w$, $\sigma^2_{b,l}=\sigma^2_b$, and $\boldsymbol{\Lambda}=\frac{\sigma^2_w}{d_{in}}\boldsymbol{I}$. There are totally $d_{in}+1+2L$ undetermined parameters $\boldsymbol{\theta}$: $\sigma^2_{w,0,j}$ ($j=1,2,\cdots,d_{in}$), $\sigma^2_{b,0}, \sigma^2_{w,l}$ ($l=1,2,\cdots,L$), and $\sigma^2_{b,l}$ ($l=1,2,\cdots,L$). These parameters (a.k.a. hyperparameters in GP regression) can be learned by using maximum likelihood estimation.

\begin{figure}[H] 
\centering
\includegraphics[width=.9\textwidth]{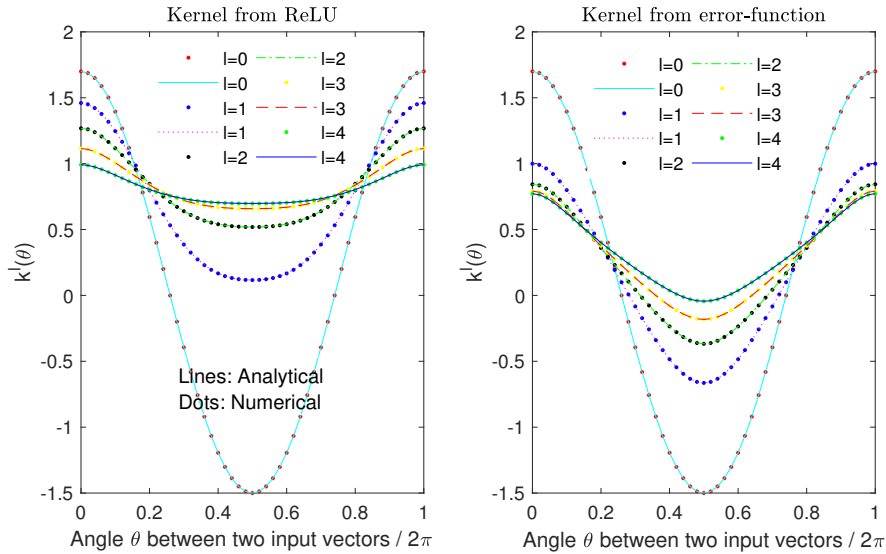}
\caption{\label{kernel_structure}Comparison of the  analytical and numerical iteration formulas for the covariance functions from the ReLU (left)  and error-function (right) nonlinearities. The covariance function $k^{l}(\boldsymbol{x},\boldsymbol{x}')$ of the GP $z^{l}_i(\boldsymbol{x})$ is computed analytically and numerically for $l=0,1,2,3,4 (L=4)$. For ease of comparison, the input vectors $\boldsymbol{x},\boldsymbol{x}'\in \mathbb{R}^2$ ($d_{in}=2$) are normalized to $||\boldsymbol{x}||_2=||\boldsymbol{x}'||_2=1$. The angle between the two vectors is $\theta=\cos^{-1}(\boldsymbol{x}^T\boldsymbol{x}')$. All the weight variances are set to be 1.6 and the bias variances are 0.1. }
\end{figure}

Since the trivariate function $F_{\phi}()$ is not analytically tractable for a generic nonlinearity, the authors of \cite{lee2017deep} developed an efficient Gaussian quadrature scheme to approximate the two-fold integral in Eq.(\ref{kl_E}). Fig.
\ref{kernel_structure} (left) compares the numerical results computed by the Gaussian quadrature scheme with the analytical results computed by Eq.  (\ref{iter_ReLU}), showing good agreement.

\subsection{Kernel from the error-function nonlinearity}
The error-function nonlinearity can also yield an analytical iteration formula just as the ReLU nonlinearity does. However, the existing formula only holds for single-hidden layer, which is given by \cite{williams1997computing}. The present paper extends the single-hidden layer case to multi-layer case using the strategy presented in Section 2.3 of \cite{cho2009kernel}.
The resulting iteration formula is 
\begin{equation}\label{iter_erf}
\begin{split}
k^{l}(\boldsymbol{x},\boldsymbol{x}')&=\frac{2\sigma^2_{w,l}}{\pi}\sin^{-1}\left(\frac{2k^{l-1}(\boldsymbol{x},\boldsymbol{x}')}{\sqrt{(1+2k^{l-1}(\boldsymbol{x},\boldsymbol{x}))(1+2k^{l-1}(\boldsymbol{x}',\boldsymbol{x}'))}}\right)+\sigma^2_{b, l},l=1,2,\cdots,L,\\
k^0(\boldsymbol{x},\boldsymbol{x}')&=\boldsymbol{x}^T\Lambda\boldsymbol{x}'+\sigma^2_{b,0}.
\end{split}
\end{equation}
To validate the above formula, we compare the analytical results from the above formula with the numerical results from the aforementioned Gaussian quadrature scheme, and as we see in Fig.\ref{kernel_structure} (right) the new iteration formula (\ref{iter_erf}) is correct.

We specifically call the GPs with the kernels that are induced iteratively by NNs ``NNGP'' for short. NNGP is actually a specific case of GP, but we still distinguish it from the standard GP involving an \textit{ad hoc} kernel (e.g., square exponential and Matern kernels). An exception is the output of the first hidden layer, $x^1_i(\boldsymbol{x})$. The output is also a GP and its covariance function is \cite{williams1997computing,rasmussen2006gaussian}
\begin{equation}\label{k_NN}
\begin{split}
k(\boldsymbol{x},\boldsymbol{x}')&=\frac{2}{\pi}\sin^{-1}\left(\frac{2k^0(\boldsymbol{x},\boldsymbol{x}')}{\sqrt{\left(1+2k^0(\boldsymbol{x},\boldsymbol{x})\right)\left(1+2k^0(\boldsymbol{x}',\boldsymbol{x}')\right)}}\right).
\end{split}
\end{equation}
We still regard the GP with this kernel as a standard GP rather than a NNGP, as the kernel $k(\boldsymbol{x},\boldsymbol{x}')$ corresponds to an incomplete, shallow NN without output layer. In Section \ref{Burgers_sec} we will compare the predictive ability of this kernel with our generic kernel $k^{l}(\boldsymbol{x},\boldsymbol{x}')$ computed by formula (\ref{iter_erf}).

\section{GP regression}\label{GPR-sec}
The essence of GP regression is the conditional distribution of quantities of interest, $\boldsymbol{q}$, given the known observations $\boldsymbol{o}$, where $\boldsymbol{q}$ and $\boldsymbol{o}$ are both Gaussian random vectors. Thanks to the analytical tractability of Gaussian distribution, we can derive analytically mean vector and covariance matrix of the conditional distribution. Supposed that $\boldsymbol{q}$ and $\boldsymbol{o}$ have the joint multivariate Gaussian distribution with zero mean vector and block-wise covariance matrix 
\begin{equation}\label{Prior}
\left[\begin{array}{c}\boldsymbol{q}\\ \boldsymbol{o}\end{array}\right]\sim
\mathcal{N}\left(\boldsymbol{0},\left[
\begin{array}{cc}\boldsymbol{K}_{qq} & \boldsymbol{K}_{qo}\\ \boldsymbol{K}_{oq} & \boldsymbol{K}_{oo}\end{array}\right]\right),
\end{equation}
the conditional distribution $p(\boldsymbol{q}|\boldsymbol{o})$ is also a multivariate Gaussian distribution with the mean vector \begin{equation}\label{post_mean}
\boldsymbol{m}(\boldsymbol{q})=\boldsymbol{K}_{qo}
\boldsymbol{K}_{oo}^{-1}\boldsymbol{o}
\end{equation}
and the covariance matrix 
\begin{equation}\label{post_cov}
\boldsymbol{K}(\boldsymbol{q})=\boldsymbol{K}_{qq}-\boldsymbol{K}_{qo}\boldsymbol{K}_{oo}^{-1}\boldsymbol{K}_{oq}
\end{equation}
(See (A.5) and (A.6) of \cite{rasmussen2006gaussian}). Zero-mean and $\sigma_{noise}^2$-variance Gaussian white noise is assumed in observations.

Note that observations and quantities of interest are usually evaluated at specific time-space coordinates. To preserve a multivariate Gaussian distribution (\ref{Prior}) at arbitrary time-space coordinates, the observations and the quantities of interest can be seen as samples of GPs. In other words, $\boldsymbol{o}=[X_o(\boldsymbol{x}_1),\cdots,X_o(\boldsymbol{x}_{N_o})]^T$ is a sample of $X_{o}(\boldsymbol{x})\sim\mathcal{GP}(0,k_o(\boldsymbol{x},\boldsymbol{x}'))$ and $\boldsymbol{q}=[X_q(\boldsymbol{x}_1),\cdots,X_q(\boldsymbol{x}_{N_q})]^T$ is a sample of $X_{q}(\boldsymbol{x})\sim\mathcal{GP}(0,k_q(\boldsymbol{x},\boldsymbol{x}'))$, where the index set $\boldsymbol{x}$ represents arbitrary time and/or space coordinates. Particularly, for solving PDEs, the observations $\boldsymbol{o}$ consist of samples from different GPs. Different covariance function $k_o(\cdot,\cdot)$ will capture different trends and features that the observations $\boldsymbol{o}$ exhibit.

Before computing mean and covariance of conditional distribution from formulas (\ref{post_mean}) and (\ref{post_cov}), we need to first learn the parameters hidden in covariance function as well as the noise variance $\sigma_{noise}^2$. These parameters are also known as hyperparameters and can be learned by using maximum likelihood estimation (MLE) \cite{rasmussen2006gaussian,le2013multi}. In practice, we find the optimal hyperparameters (including undetermined parameters $\boldsymbol{\theta}$ in covariance function and the noise variance $\sigma_{noise}^2$) that minimize the negative log marginal-likelihood \cite{rasmussen2006gaussian}
\begin{equation}
\frac{1}{2}\boldsymbol{o}^T\boldsymbol{K}^{-1}_{oo}(\boldsymbol{\theta},\sigma^2_{noise})\boldsymbol{o}+\frac{1}{2}\log|\boldsymbol{K}_{oo}(\boldsymbol{\theta},\sigma^2_{noise})|+\frac{N_o}{2}\log2\pi,
\end{equation}
where $|\cdot|$ denotes the determinant and the covariance matrix for observations (or training data) $\boldsymbol{K}_{oo}$ depends on $\boldsymbol{\theta}$ and $\sigma^2_{noise}$. 

In what follows, we will briefly review that different formulations of observations $\boldsymbol{o}$ will enable the GP regression to solve different problems including function approximation and PDE solution.

\subsection{Function approximation}
Let $\boldsymbol{q}=[f(\boldsymbol{x})]$ and $\boldsymbol{o}=[f(\boldsymbol{x}_1)+\epsilon,\cdots,f(\boldsymbol{x}_{N_o})+\epsilon]^T$ where $\epsilon$ is Gaussian white noise having variance $\sigma^2_{noise}$. The objective is to approximate the unknown scalar-valued function $f(\boldsymbol{x})$ given the observations $\boldsymbol{o}$.
After prescribing the covariance function $k_o(\cdot,\cdot)$ for computing the covariance matrix $\boldsymbol{K}_{oo}$ and then training the regression model by the MLE, we derive the function approximation by using formula (\ref{post_mean}). Moreover, the use of the variance computed from (\ref{post_cov}) yields the confidence interval. Denoting by $\boldsymbol{X}$ ($N_o$ by $d_{in}$ matrix) the collection of the evaluation points $\{\boldsymbol{x}_j\}$ (usually called training inputs), we formulate the covariance matrix in (\ref{Prior}) by:
\begin{equation}\label{cov_mat_fun}
\begin{split}
\boldsymbol{K}_{oo}&=\left[k_o(\boldsymbol{X},\boldsymbol{X})+\sigma^2_{noise}\boldsymbol{I}\right],\\
\boldsymbol{K}_{qo}&=\boldsymbol{K}_{oq}^T=\left[k_o(\boldsymbol{x},\boldsymbol{X})\right],\\
\boldsymbol{K}_{qq}&=[k_o(\boldsymbol{x},\boldsymbol{x})].
\end{split}
\end{equation}

\subsection{PDE solution}
First we consider the following time-independent linear PDE
\begin{equation}
\begin{split}
L_{\boldsymbol{x}}u(\boldsymbol{x}) & =f(\boldsymbol{x}), \boldsymbol{x}\in \Omega\in\mathbb{R}^{d_{in}},\\
u(\boldsymbol{x}) & =g(\boldsymbol{x}), \boldsymbol{x}\in \partial\Omega,
\end{split}
\end{equation}
where $L_{\boldsymbol{x}}$ is a linear differential operator. The source term $f(\cdot)$ and the boundary condition term $g(\cdot)$ are the only information we know, and thus we put them in the observations by letting $\boldsymbol{o}=[\boldsymbol{o}_u^T,\boldsymbol{o}_f^T]^T=[g(\boldsymbol{x}_{u,1})+\epsilon_u,\cdots,g(\boldsymbol{x}_{u,N_u})+\epsilon_u,f(\boldsymbol{x}_{f,1})+\epsilon_f,\cdots,f(\boldsymbol{x}_{f,N_f})+\epsilon_f]^T$, where we select $N_u$ boundary points as the evaluation points for $\boldsymbol{o}_u$ and $N_f$ domain points as the evaluation points for $\boldsymbol{o}_f$. The variances of the Gaussian white noises $\epsilon_u$ and $\epsilon_f$ are $\sigma^2_u$ and $\sigma^2_f$, respectively. We assume that the solution $\boldsymbol{q}=[u(\boldsymbol{x})]$ is a GP with zero mean and covariance function $k_u(\boldsymbol{x},\boldsymbol{x}')$, and the source term is another GP that depends on $u$, whose covariance function is written as \cite{raissi2017inferring} 
\begin{equation}\label{kernel_f}
k_f(\boldsymbol{x},\boldsymbol{x}')=L_{\boldsymbol{x}}L_{\boldsymbol{x}'}k_u(\boldsymbol{x},\boldsymbol{x}').
\end{equation}
Denoting by $\boldsymbol{X}_u$ ($N_u$ by $d_{in}$ matrix) the collection of the boundary evaluation points $\{\boldsymbol{x}_{u,j}\}$ and by $\boldsymbol{X}_f$ ($N_f$ by $d_{in}$ matrix) the collection of domain evaluation points $\{\boldsymbol{x}_{f,j}\}$, we formulate the covariance matrix in (\ref{Prior}) as \cite{raissi2017inferring}
\begin{equation}\label{cov_mat_PDE1}
\begin{split}
\boldsymbol{K}_{oo}&=\left[\begin{array}{cc}k_u(\boldsymbol{X}_u,\boldsymbol{X}_u)+\sigma^2_{noise,u}\boldsymbol{I}& L_{\boldsymbol{x}'}k_u(\boldsymbol{X}_u,\boldsymbol{X}_f)\\ L_{\boldsymbol{x}}k_u(\boldsymbol{X}_f,\boldsymbol{X}_u) & L_{\boldsymbol{x}}L_{\boldsymbol{x}'}k_u(\boldsymbol{X}_f,\boldsymbol{X}_f)+\sigma^2_{noise,f}\boldsymbol{I}\end{array}\right],\\
\boldsymbol{K}_{qo}&=\boldsymbol{K}_{oq}^T=\left[\begin{array}{cc}k_u(\boldsymbol{x},\boldsymbol{X}_u) & L_{\boldsymbol{x}'}k_u(\boldsymbol{x},\boldsymbol{X}_f)\end{array}\right],\\
\boldsymbol{K}_{qq}&=[k_u(\boldsymbol{x},\boldsymbol{x})].
\end{split}
\end{equation}

For time-dependent problem, the above GP regression model also works as we can regard the time variable as the $(d_{in}+1)$-th dimension of space-time domain (See Section 4.3.1 of \cite{raissi2017inferring}). 

The alternative way to handle the time-dependent problems is numerical GP regression \cite{raissi2017numerical}. Consider the following time-dependent PDE
\begin{equation}
\begin{split}
\frac{\partial u(\boldsymbol{x},t)}{\partial t}&=\tilde{L}_{\boldsymbol{x}}u(\boldsymbol{x},t), \boldsymbol{x}\in \Omega\in\mathbb{R}^{d_{in}},\\
u(\boldsymbol{x},0) &= h(\boldsymbol{x}),\\
u(\boldsymbol{x},t) & =g(\boldsymbol{x},t), \boldsymbol{x}\in \partial\Omega.
\end{split}
\end{equation}
The differential operator $\tilde{L}_{\boldsymbol{x}}$ can be nonlinear (say, Burgers' equation). We denote by $L_{\boldsymbol{x}}$ the linearized counterpart of $\tilde{L}_{\boldsymbol{x}}$. For linear operator, $L_{\boldsymbol{x}}=\tilde{L}_{\boldsymbol{x}}$. The use of Eular backward scheme leads to $u^n(\boldsymbol{x})-\Delta t L_{\boldsymbol{x}} u^n(\boldsymbol{x})=u^{n-1}(\boldsymbol{x})$. According to the known information from initial-boundary conditions, we define the observations at $n$-th time step as $\boldsymbol{o}^n=[\boldsymbol{o}_n^T,\boldsymbol{o}_{n-1}^T]^T=[u^n(\boldsymbol{x}_{b,1})+\epsilon_n,\cdots,u^n(\boldsymbol{x}_{b,N_b})+\epsilon_n,u^{n-1}(\boldsymbol{x}_1)+\epsilon_{n-1},\cdots,u^{n-1}(\boldsymbol{x}_{N^{n-1}})+\epsilon_{n-1}]^T$ and $\boldsymbol{q}^n=[u^n(\boldsymbol{x})]$. Note that $u^n(\boldsymbol{x}_{b,j})=g(\boldsymbol{x}_{b,j},n\Delta t)$ and $u^0(\boldsymbol{x}_j)=h(\boldsymbol{x}_j)$. Thus, we actually apply GP regression at each time step. Assuming $u^n(\boldsymbol{x})$ is a GP with zero mean and covariance kernel $k_{u^n}(\boldsymbol{x},\boldsymbol{x}')$ and denoting by $\boldsymbol{X}_b^n$ ($N_b$ by $d_{in}$ matrix) the collection of the boundary evaluation points selected at $n$-th time step and by $\boldsymbol{X}^{n-1}$ ($N^{n-1}$ by $d_{in}$ matrix) the collection of the domain evaluation points sampled at ($n-1$)-th time step, we formulate the covariance matrix as \cite{raissi2017numerical}
\begin{equation}\label{cov_mat_PDE2}
\begin{split}
\boldsymbol{K}_{o^no^n}&=\left[\begin{array}{cc}k_{u^n}(\boldsymbol{X}^n_b,\boldsymbol{X}^n_b)+\sigma^2_{noise,n}\boldsymbol{I} & (\boldsymbol{I}-\Delta t L_{\boldsymbol{x}'})k_{u^n}(\boldsymbol{X}^n_b,\boldsymbol{X}^{n-1})\\ (\boldsymbol{I}-\Delta t L_{\boldsymbol{x}})k_{u^n}(\boldsymbol{X}^{n-1},\boldsymbol{X}^n_b) & (\boldsymbol{I}-\Delta t L_{\boldsymbol{x}})(\boldsymbol{I}-\Delta t L_{\boldsymbol{x}'})k_{u^n}(\boldsymbol{X}^{n-1},\boldsymbol{X}^{n-1})+\sigma^2_{noise,n-1}\boldsymbol{I}\end{array}\right],\\
\boldsymbol{K}_{q^n o^n}&=\boldsymbol{K}_{o^n q^n}^T=\left[\begin{array}{cc}k_{u^n}(\boldsymbol{x},\boldsymbol{X}_b^n) & (\boldsymbol{I}-\Delta t L_{\boldsymbol{x}'})k_{u^n}(\boldsymbol{x},\boldsymbol{X}^{n-1})\end{array}\right],\\
\boldsymbol{K}_{q^n q^n}&=[k_{u^n}(\boldsymbol{x},\boldsymbol{x})].
\end{split}
\end{equation}

Generally, we let the covariance function be the same for all $n$, namely $k_{u^n}(\boldsymbol{x},\boldsymbol{x}')=k_o(\boldsymbol{x},\boldsymbol{x}';\boldsymbol{\theta}_n)$, except with varied undetermined parameters $\boldsymbol{\theta}_n$. Considering the propagation of the uncertainty in time marching, we need to take into account the uncertainty of the predictions in previous time step (i.e., $u^{n-1}(\boldsymbol{x})$ predicted at $\boldsymbol{X}^{n-1}$). We rewrite the predictive or posterior variance in (\ref{post_cov}) as \cite{raissi2017numerical}
\begin{equation}
\begin{split}
\boldsymbol{K}(\boldsymbol{q}^n)&=\boldsymbol{K}_{q^n q^n}-\boldsymbol{K}_{q^n o^n}\boldsymbol{K}_{o^n o^n}^{-1}\boldsymbol{K}_{o^n q^n}\\
 & +\boldsymbol{K}_{q^n o^n}\boldsymbol{K}_{o^n o^n}^{-1}
\left[\begin{array}{cc}\boldsymbol{0} & \boldsymbol{0}\\
\boldsymbol{0} & \boldsymbol{K}(u^{n-1}(\boldsymbol{X}^{n-1}))\end{array}\right]\boldsymbol{K}_{o^n o^n}^{-1}\boldsymbol{K}_{o^n q^n }.
\end{split}
\end{equation}

\section{Numerical results}\label{numerical-sec}
We perform a three-way comparison between NNGP, NN, and GP for function approximation and PDE solution for prototypical problems. The hyperparameters of NNGP and GP are trained by minimizing the negative log marginal-likelihood function. The conjugate gradient algorithm (see the function \textit{minimize}() in GPML Matlab code (version 3.6) \cite{rasmussen2010gaussian}) is employed for the optimization. For NN, network parameters (weights and biases) are trained by the L-BFGS-B algorithm that is provided in the TensorFlow package.  

Because of the non-convex nature of the objective function, to avoid trapping into the local minima, we run our optimization algorithm code ten times with different initializations. Among the ten groups of optimized hyperparameters, for GP/NNGP we choose the one yielding the smallest negative log marginal-likelihood, and for NN we select the one producing the lowest loss for NN. The initializations of hyperparameters for GP/NNGP are taken as the first ten entries of the Halton quasi-random sequence \cite{kocis1997computational} and these initializations are kept deterministic. The initializations of network weights are obtained from Xavier's initialization algorithm \cite{glorot2010understanding}, which is provided by the TensorFlow package. Also, for each initialization at most 200 function evaluations are allowed in conjugate gradient search in GP/NNGP. For NNGP, the central finite difference scheme with the step size $10^{-4}$ is used in computing the gradients of covariance function with respect to hyperparameters while for GP the analytical derivation is employed. Additionally, we compute $l_2$ relative error $||\boldsymbol{u}_{approximate}-\boldsymbol{u}_{exact}||_2/||\boldsymbol{u}_{exact}||_2$ in quantifying accuracy of function approximation and PDE solution, where $\boldsymbol{u}$ is a vector formed by the function values or PDE's solutions evaluated at test points.

\subsection{Function approximation}
The covariance matrices of GP/NNGP are formulated according to (\ref{cov_mat_fun}).

\subsubsection{Step function}
\begin{figure}[H] 
\centering
\includegraphics[width=.8\textwidth]{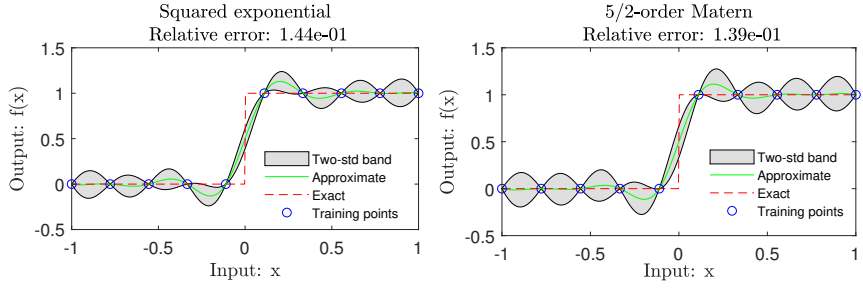}
\caption{\label{step-GP}Approximating the step function: GPs with SE (left) and Matern (right) kernels. ``Two-std band'' denotes the mean vector of the conditional distribution ($m(\boldsymbol{q})$ computed from (\ref{post_mean})) plus/minus twice the standard deviation (diagonals of $\sqrt{\boldsymbol{K}(\boldsymbol{q})}$ computed from (\ref{post_cov})). The quantities of interest is $\boldsymbol{q}=f(\boldsymbol{X}_t)$ where $\boldsymbol{X}_t$ (100 by 1 matrix) is the collection of test points.}.
\end{figure}

\begin{figure}[H] 
\centering
\includegraphics[width=.9\textwidth]{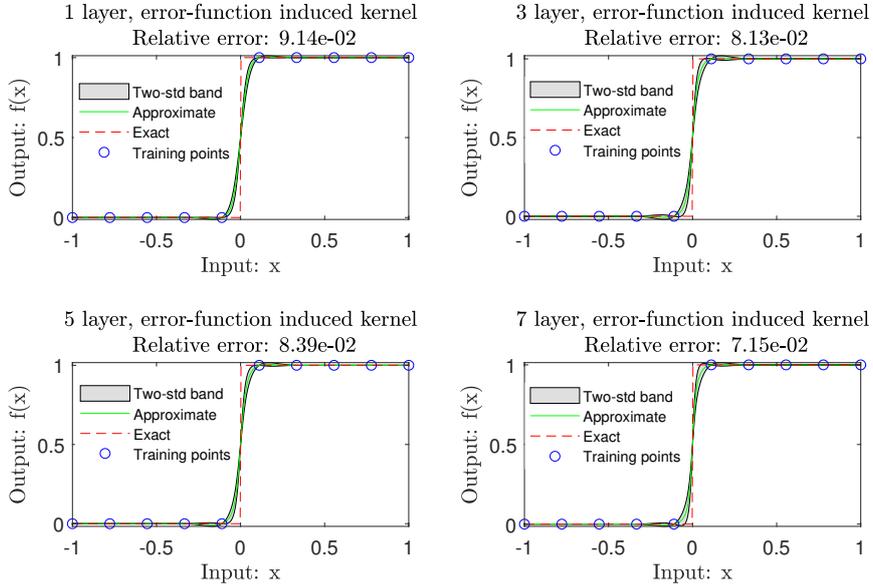}
\caption{\label{step-erf}Approximating the step function: NNGP with error-function induced kernel. ``One-layer'' indicates that there is totally one hidden layer in the inducing NN ($L=1$).}
\end{figure}

\begin{figure}[H] 
\centering
\includegraphics[width=.9\textwidth]{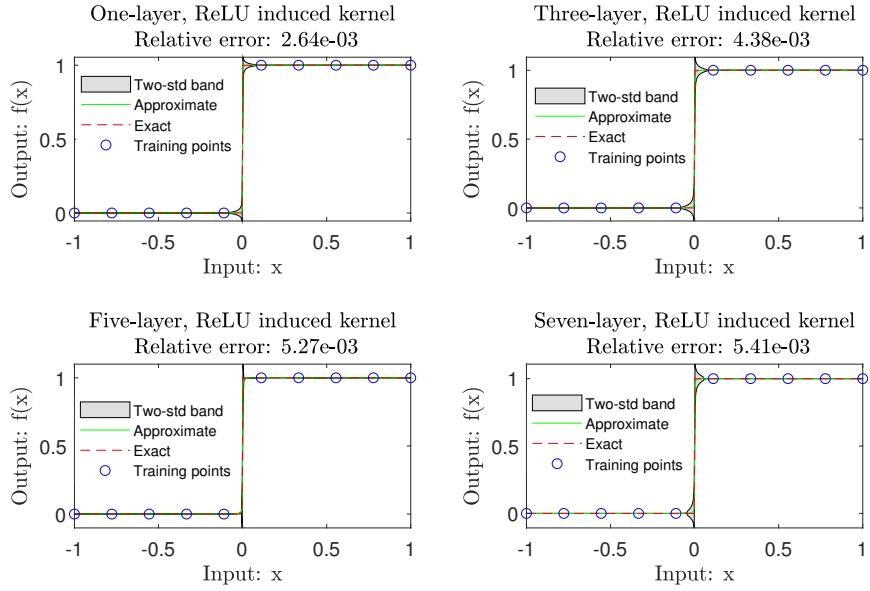}
\caption{\label{step-ReLU}Approximating the step function: NNGP with ReLU induced kernel.}
\end{figure}

\begin{figure}[H] 
\centering
\includegraphics[width=.85\textwidth]{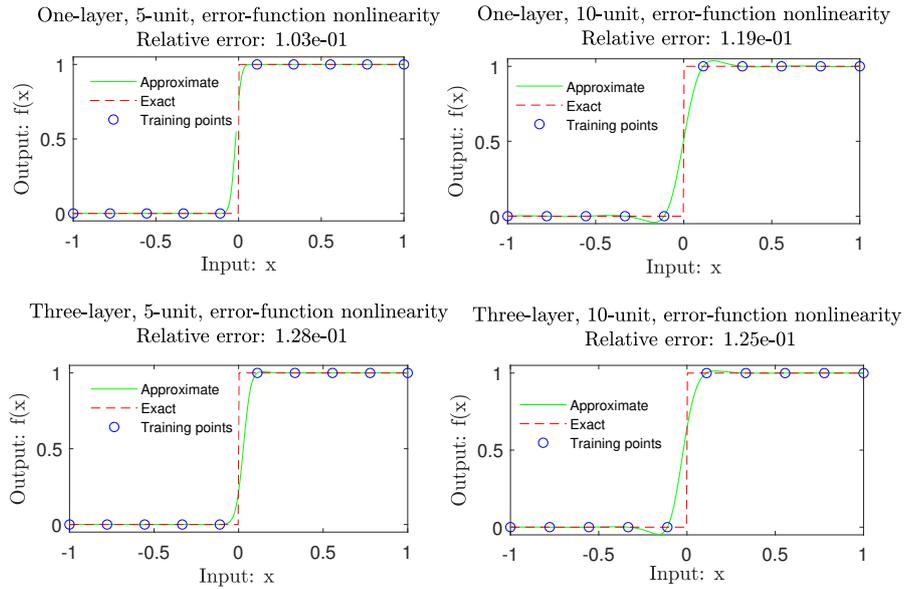}
\caption{\label{step-NN-erf}Approximating the step function: NN with the error-function nonlinearity (``5-unit'' indicates that there exist 5 units/neurons in each hidden layer).}
\end{figure}

\begin{figure}[H] 
\centering
\includegraphics[width=.9\textwidth]{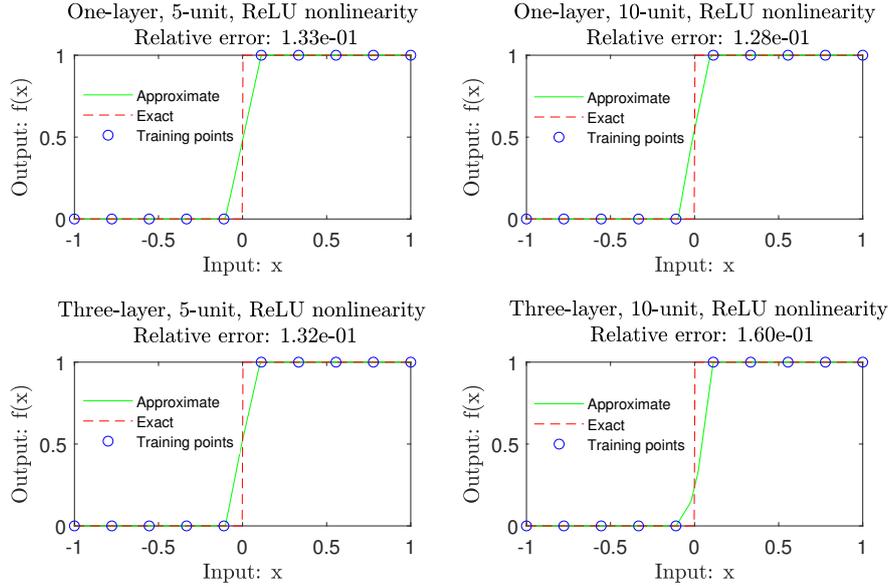}
\caption{\label{step-NN-ReLu}Approximating the step function: NN with the ReLU nonlinearity.}
\end{figure}

We approximate the step function $f(x)=1, x \ge 0; f(x)=0$, otherwise. Ten evenly distributed training inputs and 100 equispaced test points are chosen. Fig. \ref{step-GP} shows the approximation by the GPs with two commonly used kernels: squared exponential (SE) and Matern \cite{rasmussen2006gaussian}. The first kernel can describe well the data exhibiting smooth trend while the second one is suitable for finite regularity. However for the step function, a non-smooth function, the performance of these kernels is poor. The NNGP predictions shown in Figs. \ref{step-erf} and \ref{step-ReLU} are much better. In the domain away from the discontinuity, the NNGP approximation is good with small uncertainty. Particularly, the ReLU-induced kernel even succeeds in capturing the discontinuity. An important observation is that the ReLU induced kernel is more suitable for non-smooth functions than the error-function induced kernel. Another observation regarding NNGP is that deepening the inducing NN does not change the approximation accuracy for the step function. Figs. \ref{step-NN-erf} and \ref{step-NN-ReLu} show the predictions of NN, which are less accurate than NNGP's. Unlike GP/NNGP, NN does not quantify any uncertainty for approximation. 

\subsubsection{Hartmann 3D function}
The Hartmann function is frequently used for testing global optimization algorithms. Here we consider the trivariate case. The function is much smoother than the step function and thus we can expect GP to perform well. To test the approximation accuracy, we first generate $N$ ($N=100,200,500$, and $1000$) points by choosing the first $N$ entries of the Halton sequence, and then randomly permutate the points, followed by selecting the first 70\% points as training points and the remaining 30\% points as the test points.  

\begin{figure}[H] 
\centering
\includegraphics[width=.7\textwidth]{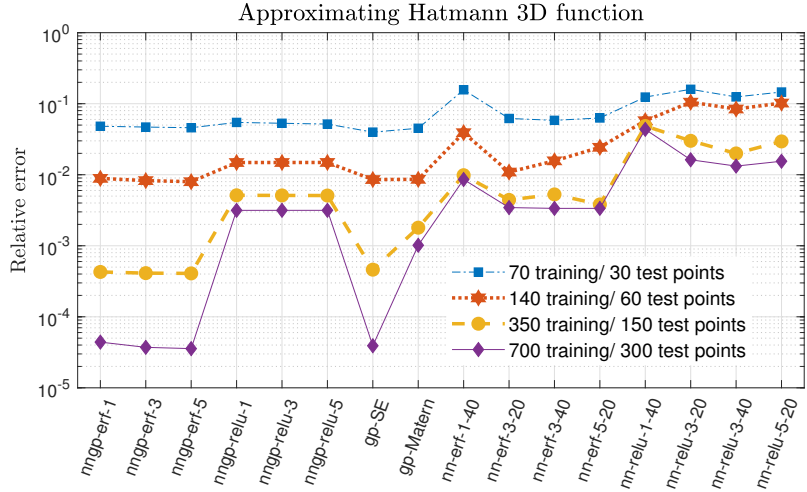}
\caption{\label{hart3-cmp-error}Approximating the Hartman 3D function: error comparison. (``nngp-erf-1'' means the NNGP with the one hidden layer and error-function induced kernel; ``gp-SE'' denotes the GP with the squared exponential kernel; ``nn-erf-3-20'' indicates the three-layer, 20-unit-wide NN with the error-function nonlinearity.)}
\end{figure}

Fig. \ref{hart3-cmp-error} compares NN, NNGP, and GP in terms of approximation accuracy. Since the Hartmann 3D function is smooth, the NNGP with error-function induced kernel and the GP with squared exponential kernel are both very accurate. ReLU is not suitable for smooth function in contrast to approximating the non-smooth step function. Additionally, the GP and NNGP both outperform NN. It is observed again that NN does not give any uncertainty estimate, whereas GP and NNGP do. Another observation is that increasing the depth of the inducing NN in NNGP does not change the accuracy.

\subsection{PDE solution}
We use the proposed covariance function from error-function nonlinearity to solve the following two PDEs. We replace the covariance function $k_u(\boldsymbol{x},\boldsymbol{x}')$ in Eq. (\ref{cov_mat_PDE1}) as well as $k_{u^n}(\boldsymbol{x},\boldsymbol{x}')$ in Eq. (\ref{cov_mat_PDE2}) with the new kernel (\ref{iter_erf}). The derivation of kernel's derivatives for the case of Poisson equation is given in Appendix.

\subsubsection{2D Poisson equation}
\begin{figure}[H] 
\centering
\includegraphics[width=.7\textwidth]{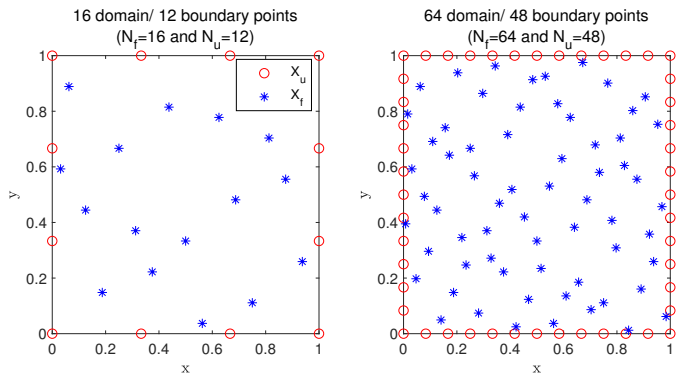}
\caption{\label{xuf}Solving 2D Poisson equation: Two setups of training inputs for both fabricated solutions.}
\end{figure}

Consider the equation
\begin{equation}
-\Delta u(x,y)=f(x,y), (x,y)\in (0,1)^2,
\end{equation}
with two fabricated solutions (1) $\sin(\pi x)(y^2+\exp(-y))$ and (2) $\sin(\pi x)\cos(2\pi(y^2+x))$. Dirichlet boundary conditions are assumed. The second solution is more complex than the first one and thus we use more training points in the second solution case. The training inputs $\boldsymbol{X}_f$ are chosen from the first $N_f$ entries of the Halton sequence and $\boldsymbol{X}_u$ is equispaced on the boundary, which are shown in Fig. \ref{xuf}.

\begin{figure}[H] 
\centering
\includegraphics[width=.7\textwidth]{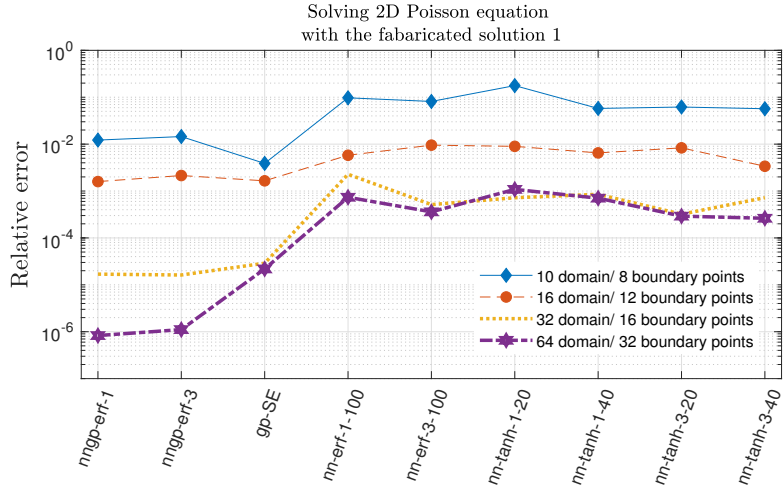}
\caption{\label{Poisson-cmp-solu1}Solving 2D Poisson equation with the exact solution $u(x,y)=\sin(\pi x)(y^2+\exp(-y))$. (``nn-tanh-3-20'' represents the three-layer, 20-unit-wide NN with the hyperbolic tangent nonlinearity.}
\end{figure}

\begin{figure}[H] 
\centering
\includegraphics[width=.7\textwidth]{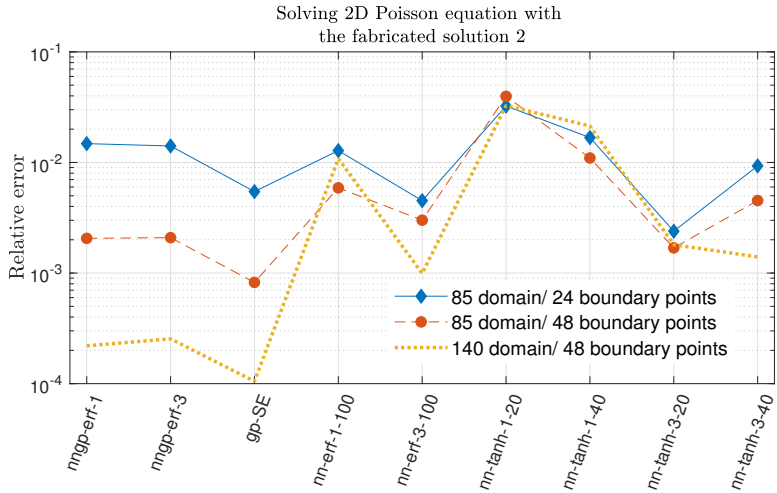}
\caption{\label{Poisson-cmp-solu2}Solving 2D Poisson equation with the exact solution $u(x,y)=\sin(\pi x)\cos(2\pi(y^2+x))$).}
\end{figure}

The covariance matrices of GP/NNGP are obtained by (\ref{cov_mat_PDE1}). Figs. \ref{Poisson-cmp-solu1} and \ref{Poisson-cmp-solu2} give the error plots of NN, NNGP, and GP for fabricated solution 1 and 2, respectively. Totally 441 evenly distributed test points are taken to evaluate the relative error. The NN results are obtained according to the approach proposed in \cite{raissi2017physics}. To keep a fair comparison, the same training inputs are selected in the NN case. It is observed from the figures that for different fabricated solutions, the approximation accuracy of the GP and NNGP is comparable. Also, the NN accuracy is nearly one order of magnitude lower than that of GP/NNGP.

\begin{figure}[H] 
\centering
\includegraphics[width=.7\textwidth]{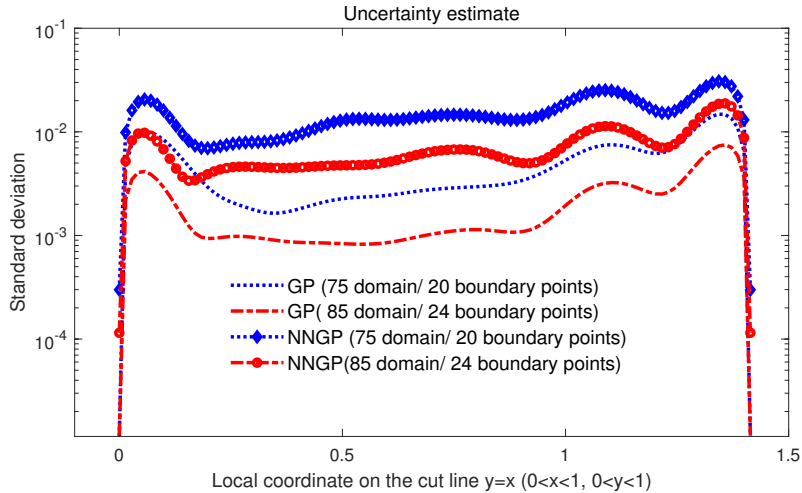}
\caption{\label{Poisson-uq-gp-nngp}Solving 2D Poisson equation with the fabricated solution 2: uncertainty estimates for GP and one-layer NNGP evaluated on the cut line $y=x$. The $y$-axis represents the standard deviation of the conditional or posterior distribution computed by (\ref{post_cov}). The relative errors evaluated on the cut line are 0.049, 0.029, 0.022, and 0.0081 from the top curve to the bottom one. It is seen that the uncertainty is strongly correlated with the prediction error. }
\end{figure}

Fig. \ref{Poisson-uq-gp-nngp} shows the uncertainty estimates of GP and one-layer NNGP evaluated on the cut line $y=x$ of the square domain, for the case of fabricated solution 2. We see that for both methods uncertainty is reduced with increasing number of training points. Moreover, the uncertainty is strongly correlated with the prediction error as smaller uncertainty corresponds to lower error. Note that the uncertainty at the endpoints of the cut line, namely, $(x,y)=(0,0)$ and $(x,y)=(1,1)$, is exactly zero, due to the boundary condition. 

\subsubsection{1D Burgers' equation}\label{Burgers_sec}
Consider the equation \cite{raissi2017numerical}
\begin{equation}
\frac{\partial u(x,t)}{\partial t}+u(x,t)\frac{\partial u(x,t)}{\partial x}=\frac{0.01}{\pi}\frac{\partial^2u(x)}{\partial x^2}, (x,t)\in [0,1]^2,
\end{equation}
with $u(-1,t)=u(1,t)=0$. The initial condition is $u(x,0)=-\sin(\pi x)$. After the linearization through replacing the nonlinear term $u^n\frac{\partial u^n}{\partial x}$ by $\mu^{n-1}\frac{\partial u^n}{\partial x}$, where $\mu^n$ is the mean vector computed for the previous time step by (\ref{post_mean}), we derive the differential operator $L_{x}=\frac{0.01}{\pi}\frac{\partial^2}{\partial x^2}-\mu^{n-1}\frac{\partial}{\partial x}$. The covariance matrices of GP/NNGP are formulated by (\ref{cov_mat_PDE2}). 

The time step size is fixed to be $\Delta t=0.01$. To evaluate the relative error at each time step, we place 400 equispaced test points in the space domain $[-1,1]$. We also solve the same equation using the NN approach proposed in \cite{raissi2017physics}. One main difference between the NN approach and the numerical GP/NNGP regression is the sampling strategy for training inputs. For NN, we sample the training inputs in the time-space domain $(x,t)\in [0,1]^2$, whereas for GP/NNGP, we sample the training inputs merely in the space domain and then perform time-marching. To make the comparison as fair as possible, we sample 10000 training points for NN, as in GP/NNGP we have at most 100 (number of time step) $\times$ 100 (number of training inputs in space) = 10000 (number of time-space sampling points). The Latin hypercube sampling strategy is adopted in sampling the training inputs in NN, GP, and NNGP. To accelerate training of numerical GP/NNGP, the initial guess of hyperparameters for current time step is taken as the optimized hyperparameters attained at previous time step. This strategy is reasonable since for a small time step two successive GPs are highly correlated and therefore the respective hyperparameters should be close. 

The kernel (\ref{k_NN}), rather than the SE considered in solving Poisson equation, is taken in GP. We choose a non-stationary kernel instead of a stationary one, because the solution exhibits discontinuity that cannot be well captured by stationary kernel. Additionally, the NN with four hidden layers of 40-unit width and the hyperbolic tangent nonlinearity is employed in the NN approach.

\begin{figure}[H] 
\centering
\includegraphics[width=.9\textwidth]{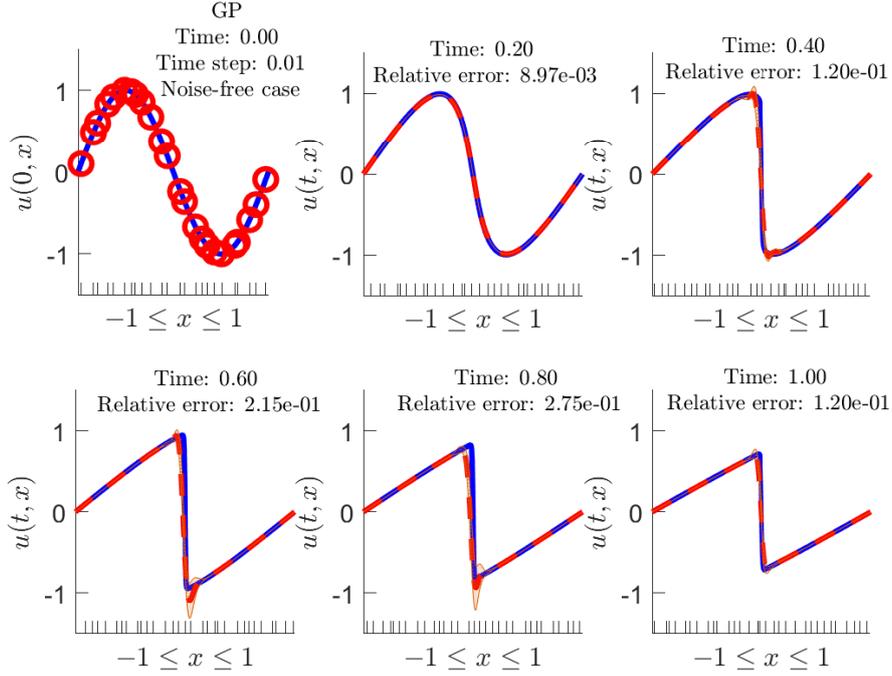}
\caption{\label{GP-100-31-noise-free}Solving 1D Burgers' equation with  noise-free initial condition: GP with kernel (\ref{k_NN}). The blue solid curve is the exact solution computed according to \cite{basdevant1986spectral}, the red circle is the training points at $n=0$ (namely $N^0=24$), and the red dashed curve is the numerical solution. The x-axis ticks denote the locations of 31 randomly sampled training points at $n\ge 1$ ($N^n=31$). The orange shadowed region is the two-standard-deviation band.}
\end{figure}

\begin{figure}[H] 
\centering
\includegraphics[width=.9\textwidth]{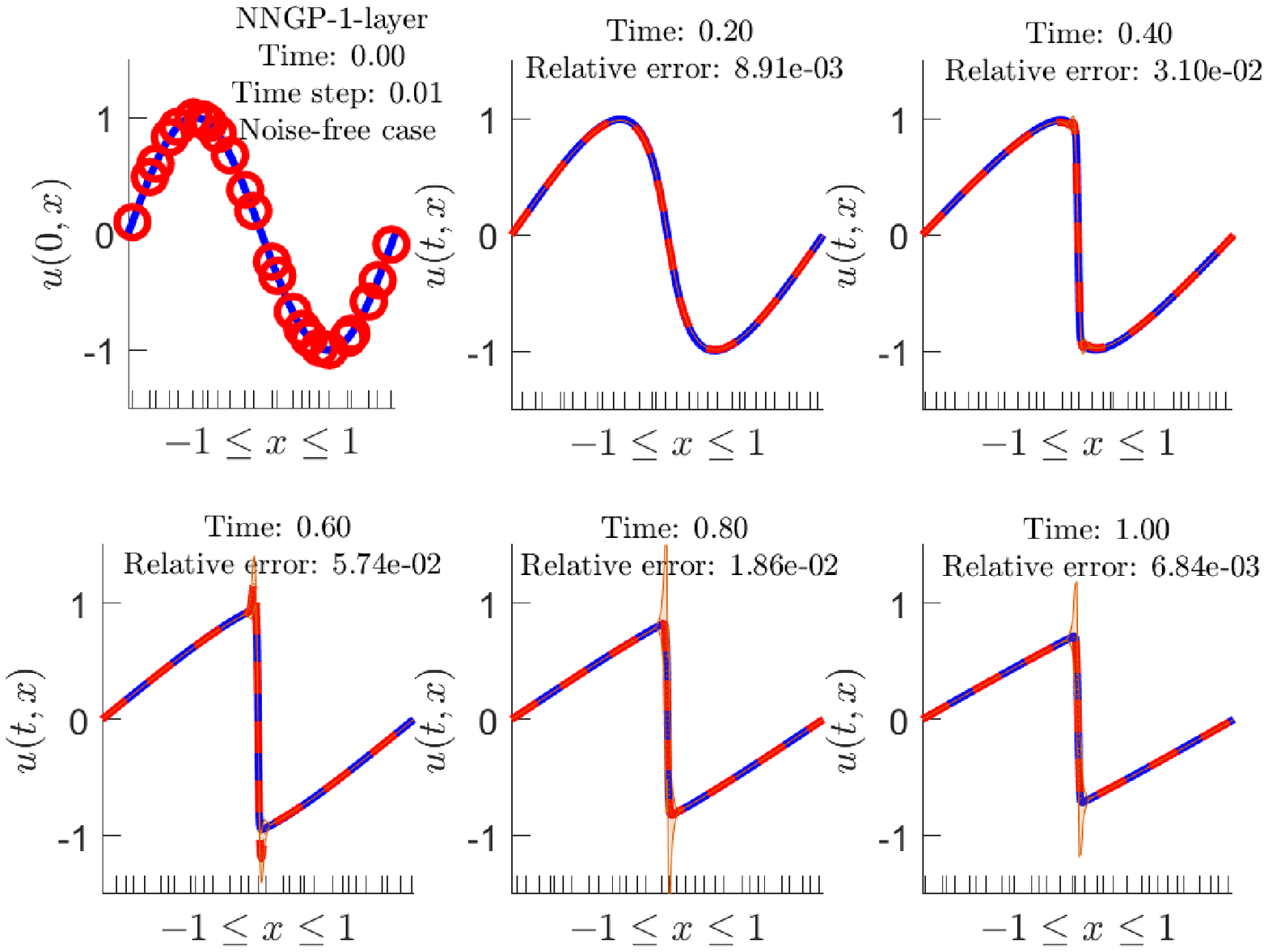}
\caption{\label{NNGP-100-31-1-noise-free}Solving 1D Burgers' equation with noise-free initial condition: NNGP with one-layer, error-function induced kernel. Number of training points at $n=0$ is $N^0=24$ and for $n\ge 1$ is $N^n=31$.}
\end{figure}

\begin{figure}[H] 
\centering
\includegraphics[width=.9\textwidth]{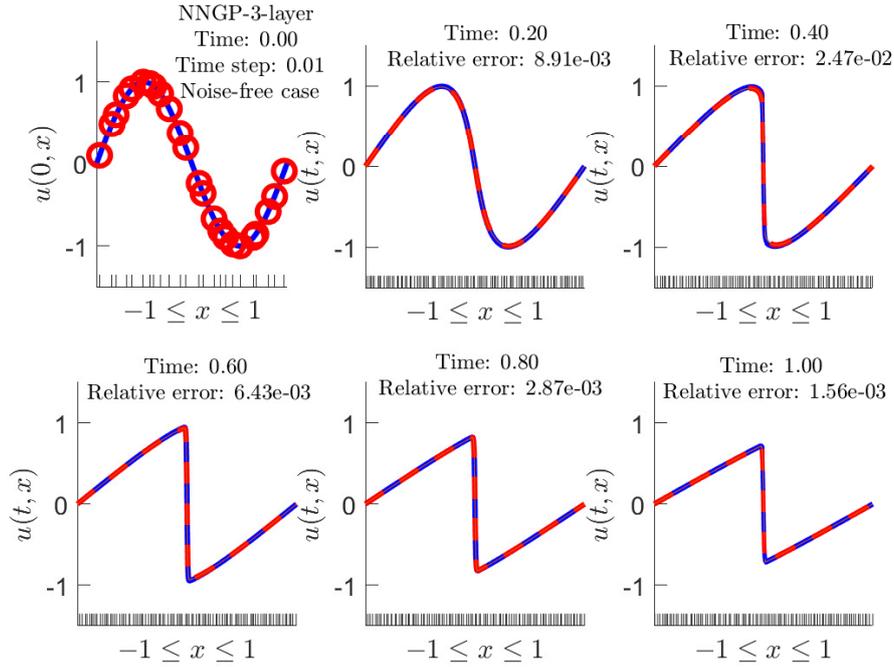}
\caption{\label{NNGP-100-101-3-noise-free}Solving 1D Burgers' equation with noise-free initial condition: NNGP with three-layer, error-function induced kernel. Number of training points at $n=0$ is $N^0=24$ and at $n\ge 1$ is $N^n=101$.}
\end{figure}

We first consider the case where the initial condition is noise-free. Figs. \ref{GP-100-31-noise-free}, \ref{NNGP-100-31-1-noise-free}, and \ref{NNGP-100-101-3-noise-free} show the numerical solutions computed by the GP, the NNGP with one-hidden-layer ($L=1$), and the NNGP with three-hidden-layer ($L=3$), respectively. We can see that the NNGP has higher solution accuracy than the GP. Importantly, different from previous examples, increasing the depth of inducing NN in NNGP improves the results in the current example. The high accuracy of NNGP is attributed to a larger number of hyperparameters compared to the GP case. More hyperparameters presumably implies higher expressivity for function approximation, which means that we can use the kernel to approximate a wider spectrum of functions. For simple function or solution (as in the Hartmann 3D function and Poisson equation examples), the advantage of higher expressivity is not fully realized, but for complex ones, such as the step function and the Burgers' equation here, the advantage can be clearly seen.

\begin{figure}[H] 
\centering
\includegraphics[width=.9\textwidth]{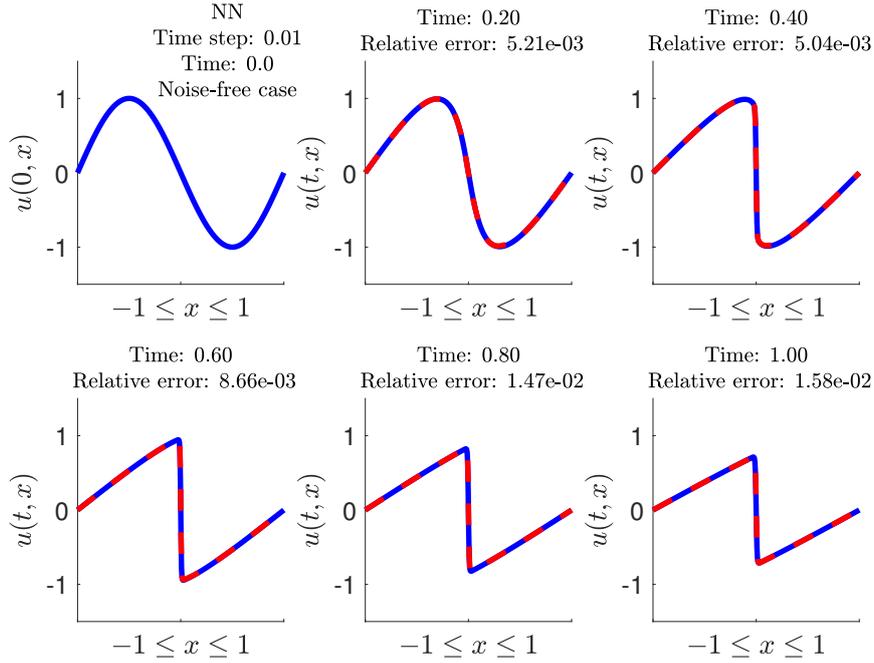}
\caption{\label{NN-tanh-4-40-noise-free}Solving 1D Burgers' equation with noise-free initial condition: NN with four hidden layers of 40-unit width as well as hyperbolic tangent nonlinearity. The maximum error appears near the discontinuity point.}
\end{figure}

\begin{figure}[H] 
\centering
\includegraphics[width=.7\textwidth]{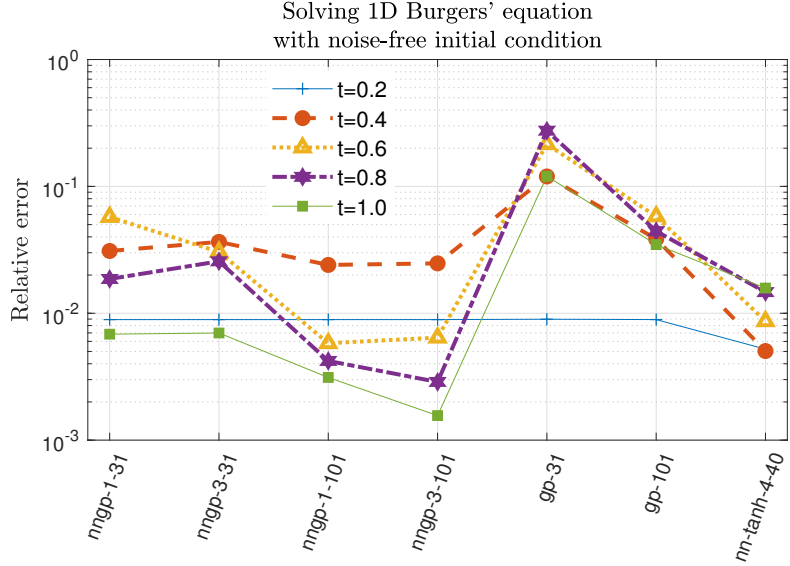}
\caption{\label{cmp-noise-free}Solving 1D Burgers' equation with noise-free initial condition: Accuracy comparison of NN, NNGP, and GP. ``nngp-1'' means the NNGP with one hidden layer and error-function induced kernel. ``31'' and ``101'' denote the numbers of training points $N^n=31$ and $101$ for $n\ge 1$, respectively. It is fixed that $N^0=24$ for all the NNGP examples. Kernel (\ref{k_NN}) is adopted in the GP. ``nn-tanh-4-40'' represents the NN with four hidden layers of 40-unit width as well as hyperbolic tangent nonlinearity.}
\end{figure}

The NN results are shown in Fig.\ref{NN-tanh-4-40-noise-free}. Although NN can derive sufficiently accurate solutions, it does not give any uncertainty estimate. Fig. \ref{cmp-noise-free} collects the errors from GP, NNGP, and NN for noise-free case. The NNGP and NN achieve errors of the same order of magnitude, while the GP performs worst. This is because NNGP and NN both have a large set of parameters to be tuned and thus possess higher expressivity.

Next we consider the case where the initial condition is contaminated by Gaussian white noise of zero mean and variance $0.15^2$. The NN approach for the noise case is beyond the scope of the present paper, since without proper regularization methods (such as dropout \cite{srivastava2014dropout}, early stopping, and weighted L1/L2 penalty), NN will easily encounter overfitting. Unlike NN, GP and NNGP inherently include the model complexity penalty in the negative log-marginal likelihood and have less risk in overfitting (see the discussion in Section 5.4.1 of the book \cite{rasmussen2006gaussian}).

\begin{figure}[H] 
\centering
\includegraphics[width=.9\textwidth]{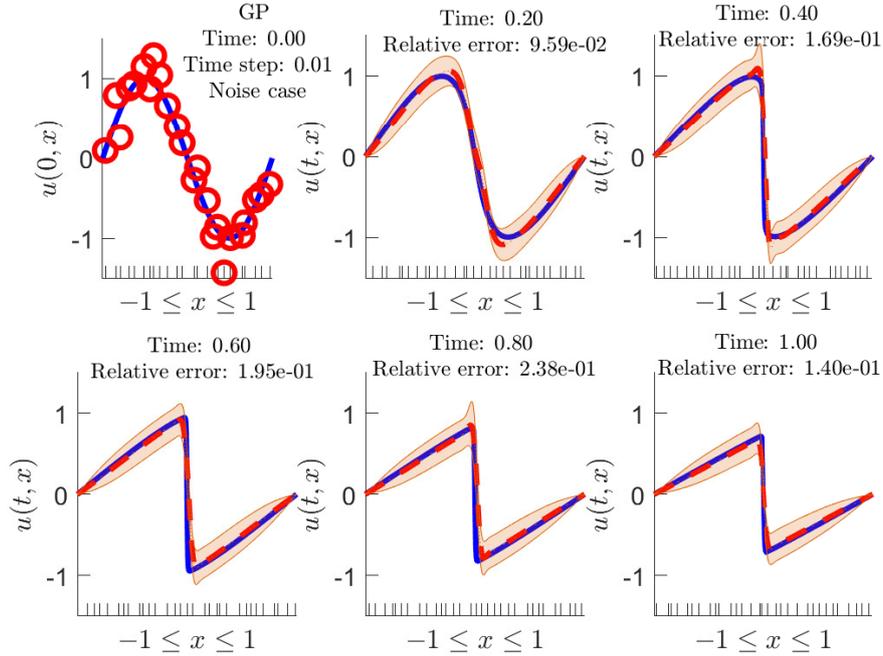}
\caption{\label{GP-100-31-noise}Solving 1D Burgers' equation with noisy initial condition (noise variance $\sigma_{noise}^2=0.15^2$): GP with kernel (\ref{k_NN}). Number of training points at $n=0$ is $N^0=24$ and at $n\ge 1$ is $N^n=31$. }
\end{figure}

\begin{figure}[H] 
\centering
\includegraphics[width=.9\textwidth]{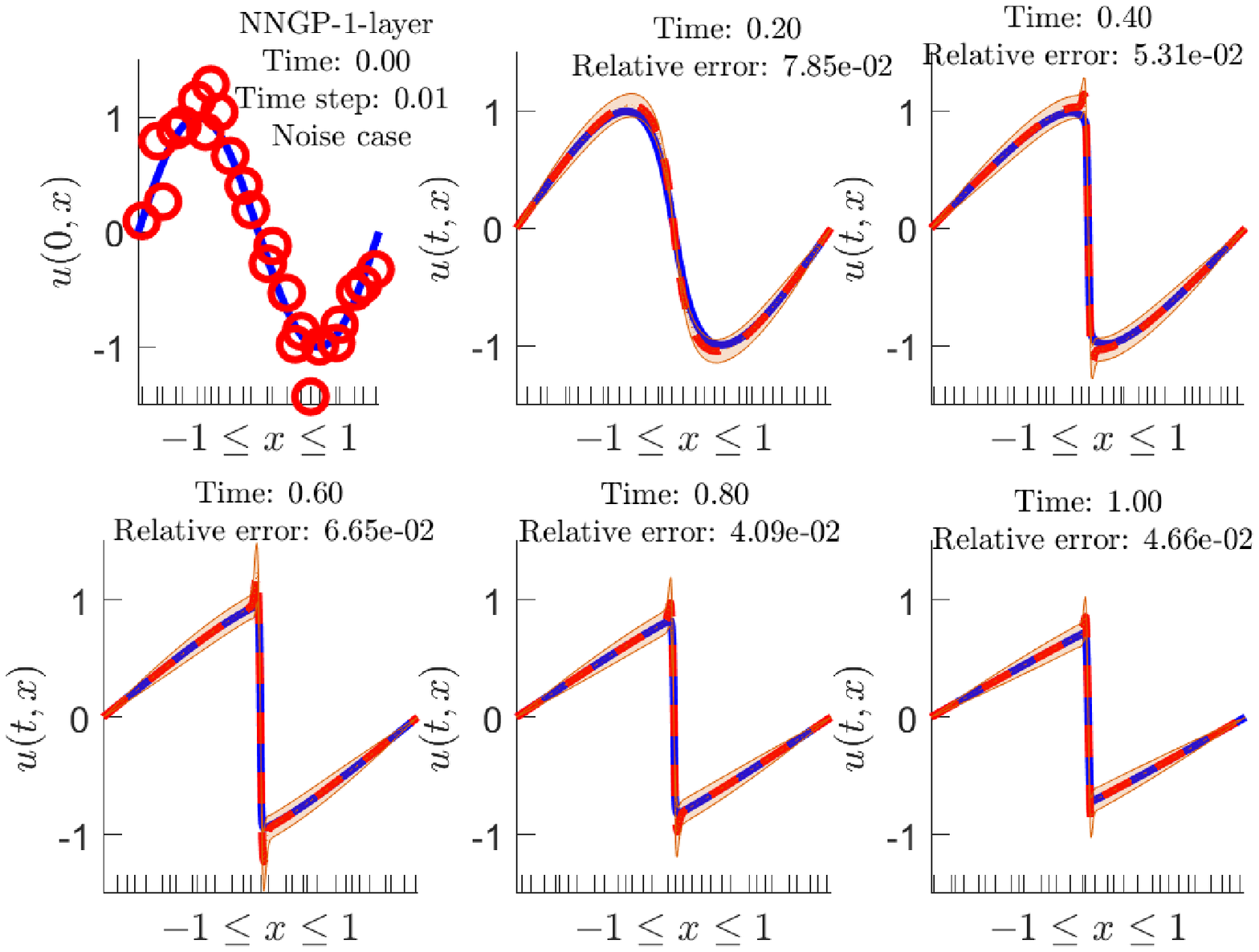}
\caption{\label{NNGP-100-31-1-noise}Solving 1D Burgers' equation with noisy initial condition (noise variance $\sigma_{noise}^2=0.15^2$): NNGP with the one hidden layer and error-function induced kernel. Number of training points at $n=0$ is $N^0=24$ and at $n\ge 1$ is $N^n=31$. }
\end{figure}

\begin{figure}[H] 
\centering
\includegraphics[width=.9\textwidth]{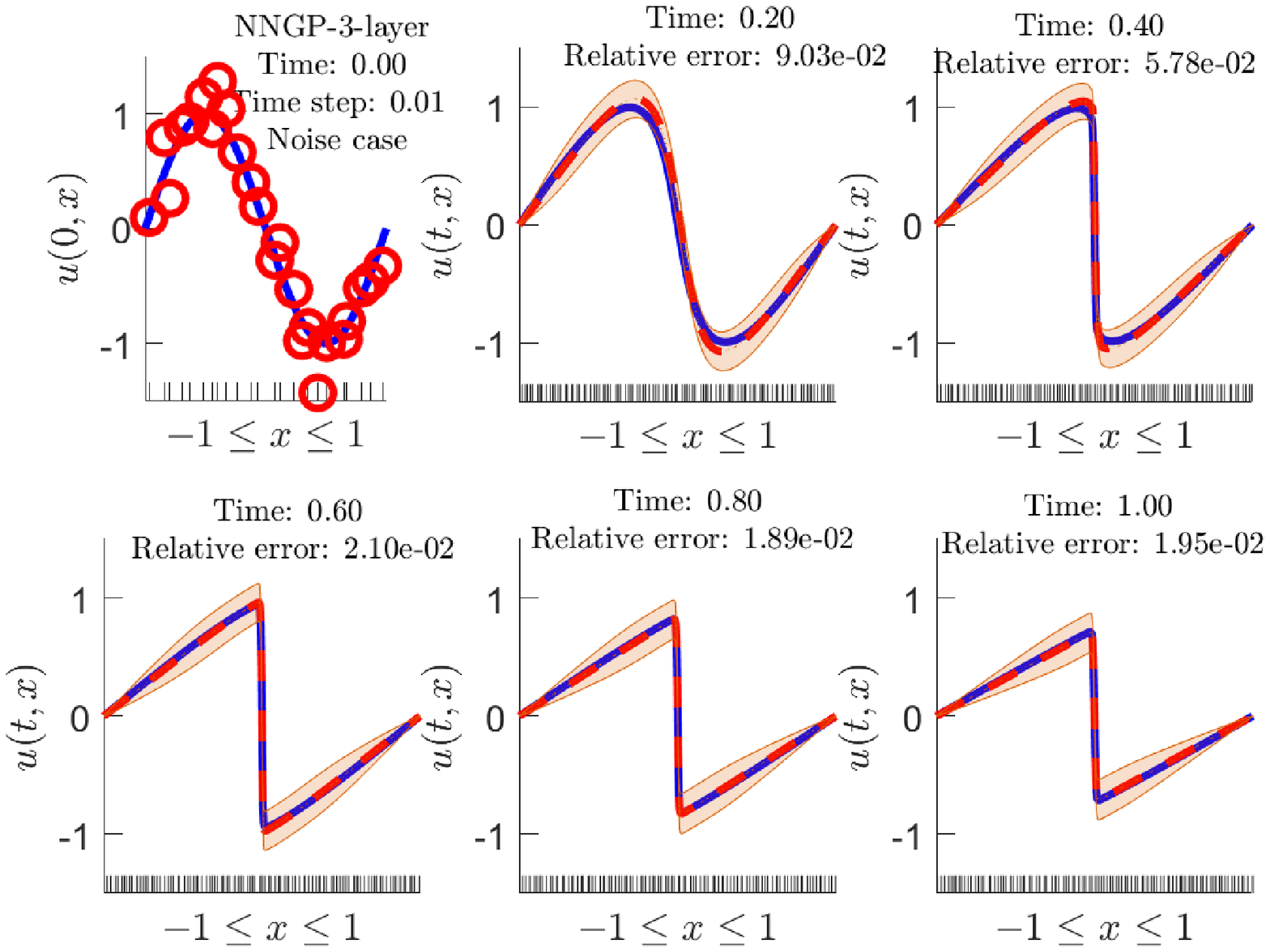}
\caption{\label{NNGP-100-101-3-noise}Solving 1D Burgers' equation with noisy initial condition (noise variance $\sigma_{noise}^2=0.15^2$): NNGP with the three hidden layers and error-function induced kernel. Number of training points at $n=0$ is $N^0=24$ and at $n\ge 1$ is $N^n=101$.}
\end{figure}

\begin{figure}[H] 
\centering
\includegraphics[width=.7\textwidth]{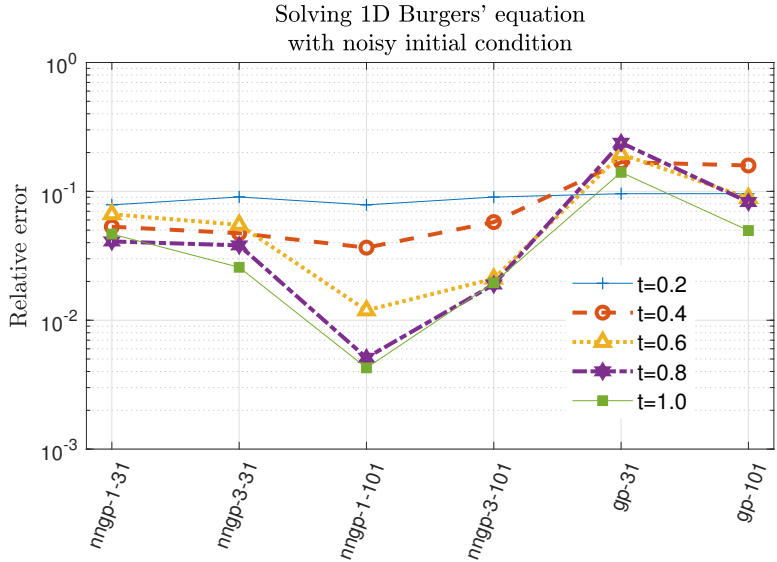}
\caption{\label{cmp-noise}Solving 1D Burgers' equation with noisy initial condition (noise variance $\sigma_n^2=0.15^2$): Accuracy comparison of GP and NNGP.  }
\end{figure}

Numerical solutions computed by the GP and NNGP are ploted in Figs. \ref{GP-100-31-noise}, \ref{NNGP-100-31-1-noise}, and \ref{NNGP-100-101-3-noise}. NNGP still has higher solution accuracy than the GP due to the higher expressivity. Data noise amplifies uncertainty represented by the orange shadowed region. GP and NNGP can handle noise well, because noise variance can be directly learned from the training data and the corresponding uncertainty is quantified by the conditional (or posterior) distribution. Fig. \ref{cmp-noise} compares the solution accuracy of GP and NNGP. For long-term simulations, the accuracy of the NNGP is roughly one order higher than that of GP. 

For NNGP, increasing the depth, $L$, of inducing NN does not guarantee the increase of accuracy. For example, one-layer NNGP with 101 training points outperforms three-layer NNGP with 101 training points, as is shown in Fig. \ref{cmp-noise}. A larger $L$ amounts to a larger number of hyperparameters, which merely increases the possibility in fitting complex functions better. Actually, $L$ can also be seen as another hyperparameter of NNGP. 

\section{Summary}

A larger number of hyperparameters enables NNGP to achieve higher or comparable accuracy for both smooth and non-smooth functions in comparison with GP, which can be seen from Table \ref{accuracy-cmp-tab}. The deep NN that induces NNGP contributes its prior variances of network parameters (weights and biases) as well as its depth to the hyperparameter list of NNGP, which endows NNGP with high expressivity. On the other hand, NNGP is able to estimate uncertainty of predictions, which is crucial to noisy-data handling and active learning \cite{brochu2010tutorial}. Unlike NN, Bayesian NN can provide uncertainty estimate. However, the conventional Bayesian NN \cite{neal2012bayesian} could be time-consuming to train due to approximation of a high-dimensional integral over weight space.   

\begin{table}[H]
\caption{Comparison of GP, NNGP, and NN (Accuracy) for function approximation and PDE solution.} \label{accuracy-cmp-tab}
\begin{center}
\begin{tabular}{c|c|c|c c c}
   \toprule
   \multirow{2}{*}{Function/Solution} & \multirow{2}{*}{Feature}& \multirow{2}{*}{Accuracy comparison} & \multicolumn{3}{c}{Kernel/nonlinearity} \\ \cline{4-6} 
 & & & GP & NNGP & NN \\ 
 \hline 
 Step function & Non-smooth & NNGP> \{GP, NN\} & SE/Matern & erf/ReLU & erf/ReLU\\
 Hartmann 3D & Smooth & \{NNGP, GP\}> NN & SE/Matern & erf/ReLU & erf/ReLU\\
 Poisson eq. & Smooth & \{NNGP, GP\}> NN & SE & erf & erf/tanh \\
 Burgers' eq. & Non-smooth & \{NNGP, NN\}> GP & kernel (\ref{k_NN}) & erf & tanh \\
 \bottomrule
\end{tabular}
\end{center}
\end{table}

Due to the need for inverting the covariance matrix, NNGP has cubic time complexity for training (see Table \ref{cmp2}). In Table \ref{cmp2}, for GP and NNGP $N_{training}$ is the size of training set, and $m_{GP}$ and $m_{NNGP}$ are numbers of evaluations of negative log marginal-likelihood in conjugate gradient algorithm for GP and NNGP, respectively. For NN, the accurate estimate of time complexity for training is still an open question \cite{song2017NNcomplexity}. Generally, the training of a fully-connected NN is faster than that of GP, because one does not need to invert a matrix. For each training point, the forward and backward propagation only need linear cost, namely $O(N_{weight})$, where $N_{weight}$ is the total number of weights in the network. 
$N_{training}$ for NN means batch size; in this paper we fed the whole training set to optimization algorithm and thus the batch size is exactly the size of training set. $m_{NN}$ is number of iterations in optimization algorithm. In numerical examples of this paper, training set size does not exceed 1000. However, for very large datasets, GP and NNGP will be less attractive compared to NN. In future work, we intend to leverage scalable GPs recently developed in \cite{hensman2013gaussian,ambikasaran2016fast,litvinenko2017likelihood} to tackle large dataset problems.

\begin{table}[H]
\caption{Comparison of GP, NNGP, and NN (whether to estimate uncertainty and the computational cost for training) for function approximation and PDE solution.} \label{cmp2}
\begin{center}
\begin{tabular}{cccc}
\toprule
  & GP & NNGP  & NN \\
  \hline
  Uncertainty & \checkmark & \checkmark & $\times$ \\
  Cost & $O(m_{GP}N_{training}^3)$ & $O(m_{NNGP}N_{training}^3)$ & $O(m_{NN}N_{weight}N_{training})$ \\
  \bottomrule
 \end{tabular}
 \end{center}
 \end{table}

\section*{Acknowledgements}
We thank Mr. Dongkun Zhang and Dr. Maziar Raissi for useful discussions. This  work  was  supported  by AFOSR (FA9550-17-1-0013) and NSF (DMS-1736088). The first author was also supported by the National Natural Science Foundation of China (11701025). The second author was also supported by a gift from Swiss company BH-Robotics.

\bibliographystyle{unsrt}
\bibliography{ref.bib}

\cleardoublepage

\setcounter{equation}{0}
\renewcommand{\theequation}{A\arabic{equation}}

\begin{appendices}\label{appendix}

\section*{Appendix: Derivatives of NNGP kernel from the error-function nonlinearity}
Absorbing the constant coefficients $\frac{2}{\pi}$ and $2$ into the variances, the kernel derived from the error-function nonlinearity, namely (\ref{iter_erf}), can be further simplified to 
\begin{equation}
\begin{split}
k^{l}(\boldsymbol{x},\boldsymbol{x}')&=\sigma^2_{w,l}\sin^{-1}\left(\frac{k^l(\boldsymbol{x},\boldsymbol{x}')}{\sqrt{(1+k^{l-1}(\boldsymbol{x},\boldsymbol{x}))(1+k^{l-1}(\boldsymbol{x}',\boldsymbol{x}'))}}\right)+\sigma^2_{b, l},l=1,2,\cdots,L,\\
k^0(\boldsymbol{x},\boldsymbol{x}')&=\boldsymbol{x}^T\Lambda\boldsymbol{x}'+\sigma^2_{b,0}.
\end{split}
\end{equation}
To solve PDEs, we need to compute the derivatives of the kernel $k^{l}(\boldsymbol{x},\boldsymbol{x}')$. We take the 2D Poisson equation as an example. We need to know the explicit forms of $-\left(\frac{\partial^2}{\partial x^2}+\frac{\partial^2}{\partial y^2}\right)k^{l}(\boldsymbol{x},\boldsymbol{x}')$ and $\left(\frac{\partial^2}{\partial x^2}+\frac{\partial^2}{\partial y^2}\right)\left(\frac{\partial^2}{\partial^2x'}+\frac{\partial^2}{\partial^2y'}\right)k^{l}(\boldsymbol{x},\boldsymbol{x}')$ where $\boldsymbol{x}=[x,y]^T$ and $\boldsymbol{x}'=[x',y']^T$. The partial derivatives up to fourth order needs to be derived. Denoting by the trivariate function $F_{\phi}()$ the $\arcsin()$ term in the above iteration formula, the use of chain rule yields
\begin{equation}
\begin{split}
\frac{\partial k^{l}(\boldsymbol{x},\boldsymbol{x}')}{\partial x} &= \sigma^2_{w,l}\left[\frac{\partial F_{\phi}}{\partial k^{l-1}(\boldsymbol{x},\boldsymbol{x})}\frac{\partial k^{l-1}(\boldsymbol{x},\boldsymbol{x})}{\partial x}+\frac{\partial F_{\phi}}{\partial k^{l-1}(\boldsymbol{x},\boldsymbol{x}')}\frac{\partial k^{l-1}(\boldsymbol{x},\boldsymbol{x}')}{\partial x}+\frac{\partial F_{\phi}}{\partial k^{l-1}(\boldsymbol{x}',\boldsymbol{x}')}\times 0\right],\\
\frac{\partial^2k^{l}(\boldsymbol{x},\boldsymbol{x}')}{\partial x^2} &= \sigma^2_{w,l}\left[\left( \frac{\partial^2F_{\phi}}{\partial^2k^{l-1}(\boldsymbol{x},\boldsymbol{x})}\frac{\partial k^{l-1}(\boldsymbol{x},\boldsymbol{x})}{\partial x}+\frac{\partial^2F_{\phi}}{\partial k^{l-1}(\boldsymbol{x},\boldsymbol{x}')\partial k^{l-1}(\boldsymbol{x},\boldsymbol{x})}\frac{\partial k^{l-1}(\boldsymbol{x},\boldsymbol{x}')}{\partial x}
\right) \frac{\partial k^{l-1}(\boldsymbol{x},\boldsymbol{x})}{\partial x}
\right.\\
 & \left. + \frac{\partial F_{\phi}}{\partial k^{l-1}(\boldsymbol{x},\boldsymbol{x})}\frac{\partial^2k^{l-1}(\boldsymbol{x},\boldsymbol{x})}{\partial x^2}  \right. \\
 & \left. +\left( \frac{\partial^2F_{\phi}}{\partial k^{l-1}(\boldsymbol{x},\boldsymbol{x})\partial k^{l-1}(\boldsymbol{x},\boldsymbol{x}')}\frac{\partial k^{l-1}(\boldsymbol{x},\boldsymbol{x})}{\partial x}+\frac{\partial^2F_{\phi}}{\partial^2 k^{l-1}(\boldsymbol{x},\boldsymbol{x}')}\frac{\partial k^{l-1}(\boldsymbol{x},\boldsymbol{x}')}{\partial x}
\right) \frac{\partial k^{l-1}(\boldsymbol{x},\boldsymbol{x}')}{\partial x}\right.\\
&\left. +\frac{\partial F_{\phi}}{\partial k^{l-1}(\boldsymbol{x},\boldsymbol{x}')}\frac{\partial^2k^{l-1}(\boldsymbol{x},\boldsymbol{x}')}{\partial x^2}\right]\\
& \cdots \cdots \cdots \cdots \\
\end{split}
\end{equation}
Generally, the iteration formulas for kernel and its derivatives can be written as
\begin{equation}
\frac{\partial^{i+j}k^{l}(\boldsymbol{x},\boldsymbol{x}')}{\partial x^i\partial x'^j} = F_{i,j}\left(\frac{\partial^{i'+j'}k^{l-1}(\boldsymbol{x},\boldsymbol{x}')}{\partial x^{i'}\partial x'^{j'}},\frac{\partial^{i'+j'}k^{l-1}(\boldsymbol{x},\boldsymbol{x})}{\partial x^{i'}\partial x'^{j'}},\frac{\partial^{i'+j'}k^{l-1}(\boldsymbol{x}',\boldsymbol{x}')}{\partial x^{i'}\partial x'^{j'}},0\le i' \le i, 0\le j'\le j\right), 
\end{equation}
where $i,j \in {0,1,2}$ and $F_{0,0}=F_{\phi}$. The initial values for the iteration are the known kernel $k^0$ and its derivatives. It should be noted that the explicit form of the function $F_{i,j}$ will become rather lengthy and ugly for higher order derivatives, but fortunately, we have derived for the readers these lengthy formulas using Maple. The Matlab code computing the covariance functions and its partial derivatives up to fourth order can be downloaded at \url{https://github.com/Pang1987/nngp_kernel_derivative}.

\end{appendices}

\end{document}